\documentclass[sigconf]{aamas} 
\usepackage[utf8]{inputenc}
\usepackage[ruled,longend]{algorithm2e}
\usepackage{subcaption}
\usepackage{graphicx}
\usepackage{caption}
\usepackage{float}
\usepackage{balance} % for balancing columns on the final page

\usepackage{subfiles}
\usepackage{tikz}
\usepackage{tikz-cd}
\usepackage{hyperref}
\setcopyright{ifaamas}
\acmConference[AAMAS '23]{Proc.\@ of the 22nd International Conference
on Autonomous Agents and Multiagent Systems (AAMAS 2023)}{May 29 -- June 2, 2023}
{London, United Kingdom}{A.~Ricci, W.~Yeoh, N.~Agmon, B.~An (eds.)}
\copyrightyear{2023}
\acmYear{2023}
\acmDOI{}
\acmPrice{}
\acmISBN{}

\title[] {Controlled Diversity with Preference : Towards Learning a Diverse Set of Desired Skills}

\author{Maxence Hussonnois}
\affiliation{
  \institution{$A^2 I^2$, Deakin University}
  \city{Geelong}
  \country{Australia}}
\email{m.hussonnois@deakin.edu.au}

\author{Thommen George Karimpanal}
\affiliation{
  \institution{$A^2 I^2$, Deakin University}
  \city{Geelong}
  \country{Australia}}
\email{thommen.karimpanalgeorge@deakin.edu.au}

\author{Santu Rana}
\affiliation{
  \institution{$A^2 I^2$, Deakin University}
  \city{Geelong}
  \country{Australia}}
\email{santu.rana@deakin.edu.au}
\begin{abstract}

Autonomously learning diverse behaviors without an extrinsic reward signal has been a problem of interest in reinforcement learning. 
However, the nature of learning in such mechanisms is unconstrained, often resulting in the accumulation of several unusable, unsafe or misaligned skills. In order to avoid such issues and ensure the discovery of safe and human-aligned skills, it is necessary to incorporate humans into the unsupervised training process, which remains a largely unexplored research area.
In this work, we propose \emph{Controlled Diversity with Preference} (CDP)\footnote{See code here: \href{https://github.com/HussonnoisMaxence/CDP}(https://github.com/HussonnoisMaxence/CDP)} , a novel, collaborative human-guided mechanism for an agent to learn a set of skills that is diverse as well as desirable. The key principle is to restrict the discovery of skills to those regions that are deemed to be desirable as per a preference model trained using human preference labels on trajectory pairs. We evaluate our approach on 2D navigation and Mujoco environments and demonstrate the ability to discover diverse, yet desirable skills.

\end{abstract}


\newcommand{\BibTeX}{\rm B\kern-.05em{\sc i\kern-.025em b}\kern-.08em\TeX}

\begin{document}

%%% The following commands remove the headers in your paper. For final 
%%% papers, these will be inserted during the pagination process.

\pagestyle{fancy}
\fancyhead{}

%%% The next command prints the information defined in the preamble.

\maketitle 

%%%%%%%%%%%%%%%%%%%%%%%%%%%%%%%%%%%%%%%%%%%%%%%%%%%%%%%%%%%%%%%%%%%%%%%%

\section{Introduction}

Deep Reinforcement learning \cite{Mnih2015} is a powerful computational approach for solving sequential decision making tasks by maximizing prespecified rewards over time. Despite its proven success in a number of applications ranging from Atari games to robotics \cite{Mnih2015,Lillicrap2016}, the framework is typically task-specific, and the effectiveness of the learned policy is contingent on a carefully designed extrinsic reward function.

However, in the real world, an agent is likely to come across complex and unstructured tasks, for which it may need to learn several sub-behaviors or skills, possibly, without access to any extrinsic rewards. In order to autonomously discover and learn these skills, prior works have proposed information theory-based diversity objectives as an intrinsic reward to explore and learn diverse task-agnostic skills without a reward function \cite{Eysenbach2018, Sharma2020Dynamics-Aware, Campos2020ExploreDA}. 
\begin{figure}[ht]
  \centering
  \includegraphics[width=0.5\linewidth]{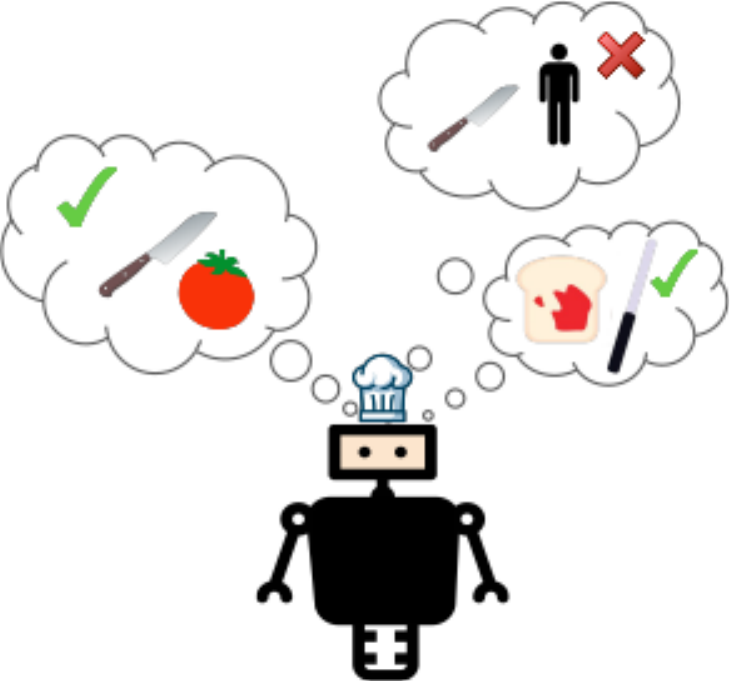} 
  \caption{With unconstrained skill discovery, a cooking robot may discover undesirable skills (such as harming humans) using a kitchen knife.}
  \label{fig:illustration}
  \Description{A cooking robot thinking about the skills that it learned with a knife, cutting vegetable, menacing a human and spreading jam on bread}
\end{figure}

While such unsupervised methods of skill discovery can produce promising results, their unconstrained nature may lead to the acquisition of useless, dangerous, or misaligned skills. For example, as depicted in Figure \ref{fig:illustration}, a robot tasked with learning diverse skills with a kitchen knife may learn undesirable skills such as harming a human. This type of behavior can occur because the agent lacks context about the real world. Without context, the agent views all aspects of the environment as equally relevant, and learns to correlate its skills with any part of the environment regardless of its importance or safety.

In order to address this issue, \citet{Eysenbach2018} attempted to limit the diversity of skills by manually selecting features for the agent to be diverse in. However, the effectiveness of this approach is limited, as it is still possible for agents to learn undesirable skills while being diverse about a specific feature. Recent works \cite{klemsdal2021learning} have suggested relying on expert demonstrations to guide the agent towards expert-visited regions. 
Such demonstrations are generally expensive and thus would not be available in large quantities, thereby negatively impacting skill diversity.
As such, designing online approaches for learning simultaneously diverse and desirable behaviors remains an important and challenging open problem. 

In contrast to the approaches mentioned above, in this work, we contend that the agent can learn more desirable skills through guidance provided by humans in the loop during the learning of skills. Using human feedback to infer context allows much greater flexibility, adaptability and less engineering than relying on predefined extrinsic rewards. The key idea behind our approach is that we frame the problem of \emph{controlling skill diversity} as finding regions of the environment where skill discovery will more likely produce desirable skills.
 Due to the difficulty of identifying such regions without human-provided context, we propose leveraging recent work in learning from human preferences \cite{wilson2012, Christiano2017, 2021pebble} to infer preferred regions in the environment. Intuitively, these are regions of the environment which are generally associated with favorable agent behaviors. We posit that such regions also correspond to suitable regions for learning a diverse set of skills. Once such regions are identified, we adapt recent exploration methods to direct the agent's exploration towards those preferred regions. Furthermore, by learning a representation of the state space from human preferences, we show that our approach scales to higher dimensional problems and learns skills that are discernibly diverse to human eyes. Thus, by restricting the diversity in skill discovery to human-preferred regions of the environment, we are capable of learning skills that are both diverse and desirable.

In summary, the main contributions of this work are:
\begin{itemize}
  \item \emph{Controlled diversity with Preference} (CDP), a novel method to control diversity in skill discovery using human preferences.
  \item Demonstration that our proposed approach guides the agent's exploration towards preferred state regions.
  \item Learning a representation of the state space for skill discovery, that contains features relevant to human preferences.
  \item Qualitative and quantitative evaluation of the proposed framework, with suitable comparisons with existing baselines for learning diverse skills.

\end{itemize}

%%%%%%%%%%%%%%%%%%%%%%%%%%%%%%%%%%%%%%%%%%%%%%%%%%%%%%%%%%%%%%%%%%%%%%%%

\section{Related Work}

\textbf{Human in the loop and Preference based RL:} Human in the loop reinforcement learning (HIL-RL) aims to improve reinforcement learning (RL) agents by using human knowledge. In contrast to imitation learning and inverse RL, HIL-RL uses human knowledge during the training process rather than prior to it.  

To enable the use of human feedback for more complex and challenging tasks, \citet{Christiano2017} 
 learned a reward model from human preference labels over trajectories. Such preference-based frameworks offer the advantage of relatively easy/intuitive supervision, while being sample efficient enough to quickly learn a reward function.

PEBBLE \cite{2021pebble} was an approach that further developed this framework to design a more sample- and feedback-efficient preference-based RL algorithm without any additional supervision. This was achieved by leveraging off-policy learning and utilizing unsupervised pre-training to collect data to substantially improve efficiency. Although we follow a PEBBLE-like approach to learn a reward function from preference labels, our approach differs from PEBBLE in that in addition to learning reward functions from preferences, we use this learned reward function to determine a distribution of states for guiding the agent's exploration. We essentially utilize the learned preference based rewards as a means for determining the dynamic space that humans prefer.

In the context of our proposed approach, Skill Preferences (SkiP) \cite{wang2022skill} is a related approach that combines skill learning and human preferences. SkiP was shown to learn a reward model over skills with human preferences and used that model to extract human-aligned skills from offline data. In contrast to SkiP, our approach targets the online learning setting, where skills are still under development when we obtain preferences.

\textbf{Unsupervised Reinforcement learning and Skill discovery:} 
Unsupervised RL is an approach for autonomously learning relevant behavior in any environment based on task-agnostic intrinsic rewards. Intrinsic rewards form the basis for many agent concepts such as curiosity \cite{Pathak2017} , novelty \cite{hao2021}, and empowerment \cite{Salge2013}. In contrast to curiosity \cite{Pathak2017}, which guides exploration towards regions where predictive models perform poorly, novelty \cite{hao2021} guides exploration toward areas that are less frequently visited. In order to maximize the agent’s future potential, empowerment approaches \cite{Salge2013} direct the agent to explore regions that offer it more possible states to visit.

DIAYN \cite{Eysenbach2018}, VIC \cite{Gregor2016}, and VALOR \cite{achiam2018} suggested an empowerment objective based on mutual information. This objective was shown to enable the discovery and acquisition of a variety of skills relevant to complex locomotion. To add predictability to the set of diverse skills, DADS \cite{Sharma2020Dynamics-Aware} formulated a variation of an objective based on mutual information. EDL \cite{Campos2020ExploreDA} showed that such skill discovery methods suffer from poor exploration, and proposed to split the process into three independent phases: exploration, discovery, and learning. In our work, we use the EDL framework, thereby separating the discovery and learning of skills. However, we integrate the discovery of skills within the exploration process by using them to gather data more from the preferred regions. 

Despite various advances in the area of autonomous skill discovery, it remains challenging to learn and discover meaningful skills in high-dimensional state spaces due to the curse of dimensionality.
Many works have mitigated this problem by learning representations of the state space, to distinguish skills based on more relevant features. \citet{nieto2021} leverages self-supervised learning of state representation techniques such as contrastive techniques to learn a compact latent representation of the states. IBOL \cite{kim2021} proposes a linearization of environments that promotes more diverse and distant state transitions.
Unlike these works, we do not learn or change the representation of the state. Instead, we redirect diversity to specific regions of the state space likely associated with meaningful, desirable skills. We note that the aforementioned methods for dealing with high dimensionality remain orthogonal to our work, and could possibly be combined with our proposed framework to realise more scalable solutions. 

As far as controlling diversity using human data is concerned, the work by \citet{klemsdal2021learning} is most closely related to ours. By leveraging prior expert data, they obtain a state projection that makes expert-visited states recognizable and, consequently, encourages skills to visit them.
However, in contrast to this approach, our proposed framework does not require access to expert trajectories. It instead only assumes a finite number of human-generated preference labels based on the agent's trajectories. We contend that this type of feedback is relatively easier to collect, with minimal cognitive load on the human collaborator.

\textbf{Restraining behavior:} 
Several works have aimed at controlling the behavior of agents. For example, \citet{DeGiacomo2018FoundationsFR} introduced restraining bolts to restrain agents' behavior by offering additional rewards when logical specifications of desired actions are satisfied. In another direction, \citet{10.5555/3535850.3535854} also augments the reward of the agent to consider the future wellbeing of others and thus restraining its behavior to reduce negative side effects. Our work differs from these in that, through interaction using human preferences, we learn how to regulate the diversity of skills.

%%%%%%%%%%%%%%%%%%%%%%%%%%%%%%%%%%%%%%%%%%%%%%%%%%%%%%%%%%%%%%%%%%%%%%%%
\section{Preliminaries}

In this paper, we consider the problem of controlling diverse skill discovery by combining the EDL framework with human guidance in the form of preference-based RL. Here, we briefly present related concepts, before describing our method in detail in Section \ref{sec:methods}.
\subsection{Skill Discovery}
Consistent with prior work \cite{Campos2020ExploreDA} the skill discovery problem is formalized as a Markov Decision Process (MDP) $\mathcal M = (\mathcal S, \mathcal A, \mathcal P)$ without external rewards, $\mathcal S$ and $\mathcal A$ respectively denote the state and action spaces, and $\mathcal P$ is the transition function. Skills  introduced by \citet{SUTTON1999181} are temporally extended actions (sub-behavior), which consist of a sequence of primitive actions.
We define skills as policies $\pi(a|s,z)$ conditioned on a fixed latent variable $z \in  Z $ .  

Skill discovery methods aim to learn these latent-conditioned policies by maximising the mutual information between $\mathcal S$ and $Z$. Due to symmetry, the corresponding mutual information can be expressed in two forms:
\begin{eqnarray}
\label{eq:mi}
I(\mathcal S;Z)= \underbrace{H(Z) - H(Z|\mathcal S)}_{\text{reverse}} =   \underbrace{H(\mathcal S) - H(\mathcal S|Z)}_{\text{forward}}  
\end{eqnarray}
where $I(\cdot;\cdot)$ and $H(\cdot)$ are respectively the mutual information and the Shannon entropy. Following prior work, we refer to these as the reverse and forward forms. By using either of the two forms of the objective, prior works \cite{Eysenbach2018,Sharma2020Dynamics-Aware,Campos2020ExploreDA} have demonstrated the learning of latent-conditioned policies that execute diverse skills. Our method uses the forward form in Equation \eqref{eq:mi} to learn the latent-conditioned policies $\pi(a|s,z)$  . 

\subsection{EDL Framework}
EDL optimizes the same information theoretic objective in Equation \eqref{eq:mi},  but separates the skill discovery into three distinct stages - exploration, discovery and learning of the skill. 
\subsubsection{Exploration stage} The Exploration stage aims to collect environment transitions; it can be achieved via any exploration method \cite{Pathak2017, Lee2020Learning, hao2021}. 
\subsubsection{Skill Discovery stage}
\label{sec:VQSD}
Given a distribution over states $p(s)$, the Skill Discovery stage trains a Vector-Quantized VAE (VQ-VAE) to model the posterior $p(z|s)$ as an encoder $p_{\phi}(z|s)$, and $p(s|z)$ as the decoder $q_{\phi}(s|z)$. The VQ-VAE has the advantage of having a discrete bottleneck, which in our case is the categorical distribution of $p(z)$. Typically, VQ-VAEs are trained to optimize for the objective: 
\begin{eqnarray}
\label{eq:vq-vaeloss}
\mathcal{L}^\text{VQ-VAE} =  \mathbb{E}_{s\sim p(s)} [\log (q_{\phi}(s| p_{\phi}(z|s))) \nonumber \\ 
+ \lVert\text{sg}[z_e(s)] - e\rVert + \beta \lVert z_e(s) - \text{sg}[e]\rVert]
\end{eqnarray}
where $z_e(s)$ and $e$ are respectively the codebook vector and the codebook index, and $sg[.]$ is the operation `stop gradient'. For more details, we refer the reader to \citet{vandenOord2017}. 
\subsubsection{Skill Learning stage}
\label{sec:SLD}
Finally, the Skill Learning stage consists of training the latent-conditioned policies $\pi_\theta(a|s,z)$ that maximize the forward form of the mutual information (Equation \eqref{eq:mi}) between states and latent variables. The corresponding reward function is then defined by:
\begin{eqnarray}
\label{eq:edl_learning}
r(s,z)= \log q_\psi(s|z), z \sim p(z)
\end{eqnarray}
where $q_\psi(s|z)$ is given by the decoder of the VQ-VAE at the discovery stage. This reward function reinforces the policy to visit states that the decoder generates for each latent variable $z$. 
Our proposed method builds upon the EDL Framework, although we enhance it via two novel contributions: (1) an exploration phase guided by preferences, which integrates with the skill discovery phase to improve coverage relevance, and (2) a way to transform $p(s)$ into a more suitable distribution for discovering desirable skills.

\subsection{Reward Learning from Preferences}
\label{sec:PBRL}

In this work, we use preference-based RL to identify preferred regions, which are then used to constrain the diversity of learned skills. In preference-based RL, a human is presented with two trajectory segments (state-action sequences) $\sigma^i$ and $\sigma^j$, and is asked to indicate their preference $y$ for one over the other. For instance, the label $y=(1,0)$  would imply that the first segment is preferred over the second.

We follow the same framework as prior works in preference-based RL \cite{wilson2012, Christiano2017, 2021pebble}, where the aim is to model the human's internal reward function responsible for the indicated preferences. This is usually done via the Bradley-Terry model \cite{Bradley1952RankAO}, which models a preference predictor using the reward function $\hat{r}_\psi$ as follows:

\begin{eqnarray}
\label{eq:BradleyTerry}
P_\psi[\sigma^i \succ \sigma^j] = \frac{\exp({\sum_t \hat{r}_\psi (s^i_t,a^i_t))}}   {\sum_{j \in \{0,1\} } \exp {(\sum_t \hat{r}_\psi (s^j_t,a^j_t))} }
\end{eqnarray}
where $\sigma^i \succ \sigma^j$ denotes the event that the segment $\sigma^i$ is preferable to the segment $\sigma^j$. As in \citet{2021pebble}, we model the reward function as a neural network with parameters $ \psi $, which is updated by minimizing the following loss:

\begin{eqnarray}
\label{eq:BradleyTerryLoss}
\mathcal{L}^\text{Reward} = - \mathbb{E}_{(\sigma^0, \sigma^1, y)\sim \mathcal D} [y(0) \log  P_\psi[\sigma^0 \succ \sigma^1] +  \nonumber \\ 
y(1)\log P_\psi[\sigma^1 \succ \sigma^0]]
\end{eqnarray}

In the current work, the above framework is used to infer context regarding the importance of each region of the environment by estimating the human's reward function $\hat{r}_\psi$ from preferences labels $y$. Specifically, we use these rewards to identify regions associated with favorable agent behaviors. For simplicity, we choose to work with the trajectory represented as state sequences rather than state-action sequences as introduced.

%%%%%%%%%%%%%%%%%%%%%%%%%%%%%%%%%%%%%%%%%%%%%%%%%%%%%%%%%%%%%%%%%%%%%%%%
\begin{figure}[ht]
  \centering
  \includegraphics[width=0.7\linewidth]{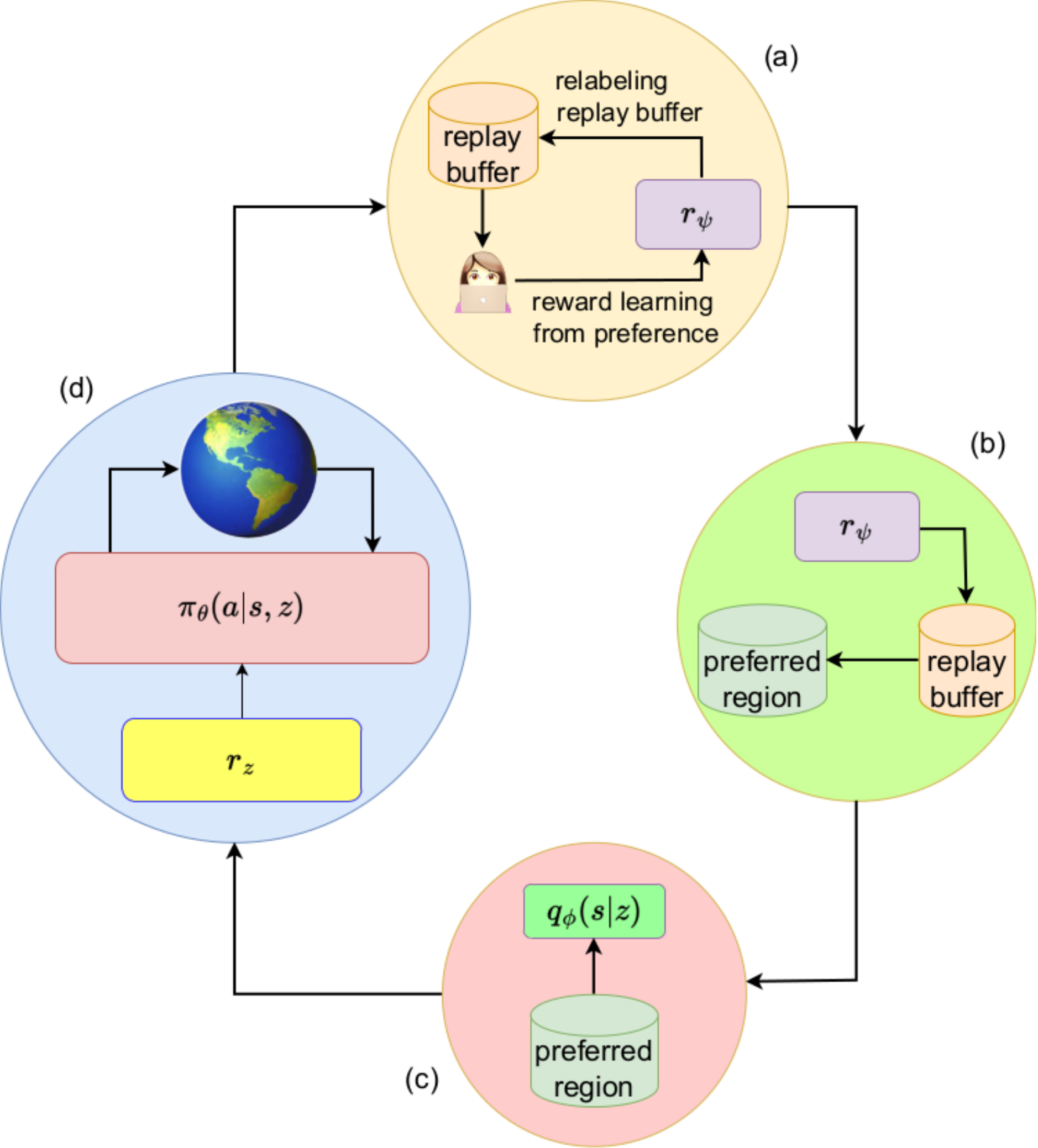} 
  \caption{Illustration of the guided exploration process. The agent iterates through four steps to explore. First 
   it learns a reward from human preferences (b) so that it can update its belief over the preferred region from the existing data in the buffer (c) then it discovers skills in this region (d) and finally it collects experience regarding the beliefs of the preferred region. }
  \label{fig:diagramme}
  \Description{Illustration of the guided exploration process. The agent iterates through four steps to explore. First it learns a reward from human preferences (b) so that it can update its belief over the preferred region from the existing data in the buffer (c) then it discovers skills in this region (d) and finally it collects experience regarding the beliefs of the preferred region.}
\end{figure}
\section {Methods}
\label{sec:methods}

In this section, we present \textbf{CDP} (\textbf{C}ontrolled \textbf{D}iversity with \textbf{P}referen
ce), 
a skill discovery method that utilizes preference-based RL methods to control diversity and discover more preferred skills based on human feedback. Our main idea is that with the reward learned from human preference feedback, we can estimate a region of the state space where it is more likely for the agent to discover desirable skills and subsequently learn them. To this end, we introduce the concept of controlled diversity and preferred regions. Then, we present how to integrate them with EDL for a more efficient exploration of the preferred region.

\subsection{Influencing Skill Discovery with Human Feedback}

\subsubsection{How to control diversity?}
We define controlled diversity as limiting diversity to a certain region of the state space. It differs from the standard setting for skill-discovery problems, where diversity is applied to the entire state space in an unconstrained manner. To achieve this, we follow the EDL framework, where we encourage skill discovery towards targeted behaviors by modifying priors through a distribution over a target region $p^*(s)$ of the state space. Performing skill discovery on $p^*(s)$ assigns latent variable $z$ to regions within the target region. In other words, a carefully designed target region containing only desirable skills will enable us to correlate $Z$ only with desirable skills.

It is however, difficult to design such a region of the state space or to gain direct access to it. Thus, we formulate our problem of `controlled diversity’ as finding an approximation of this target region. In this work, we identify such regions through their high preference rewards, learned from human preferences.

\subsubsection{Preferred regions}
\label{sec:PR}
We define a preferred region as those regions associated with high estimated preference rewards $\hat {r}_{\psi}$, where $\hat {r}_{\psi}$ is learnt using the preference based RL framework, as described in Section \ref{sec:PBRL}. Concretely, a preferred state region $\hat{S}\subseteq S$ is a region of the state space $S$ where $\hat {r}_{\psi}(s)\geq \beta$, where  $\beta \in[0,1]$ is a preference reward threshold, and $\hat {r}_\psi(s)$ is normalised to be within the range $[0,1]$. That is, 
\begin{eqnarray}
\label{eq:pref_region}
\hat S = \{\forall s \in S | \hat {r}_\psi(s) \geq \beta \}
\end{eqnarray}
Ideally, the preferred region would be aligned with the intended skills if the state space was fully explored. However, the assumption of full state coverage may not be realistic. Alternatively, we iteratively build a more accurate preference model by first using the current estimate of the preference model to sample more trajectories from the highly-preferred regions, and then updating the preference model with human labels on those trajectories. This directed sampling makes our method more query-efficient.

\subsection{Exploration Towards a Preferred Region}
\label{sec:gep}

In this section, we adapt the exploration phase of EDL to explore preferred regions more effectively. To this end, we add three components to the exploration phase - We learn a reward from human preferences, we estimate the potential preferred regions and we discover skills based on the potential preferred regions. Formally, we consider latent-conditioned policies $\pi(a|s,z)$, a
reward function $\hat{r}_{\psi}$, a preferred state region  $\hat S$  and a discriminator $q_\psi(s|z)$, which are updated by the following processes, as illustrated in Figure (\ref{fig:diagramme}): 
\begin{itemize}
  \item Step (a): Reward Learning -  We query human for preference over trajectories and update a reward function $\hat{r}_{\psi}$ from the preferences. 
  \item Step (b): Preferred regions estimations - We update our belief about the preferred subset  $\hat S$ as described in Section \ref{sec:PR}. 
  \item Step (c): Discovery - We train the discriminator $q_\psi(s|z$), following VQ-VAE training, based on the most recent belief about the preferred subset.
  \item Step (d): Exploration - We train latent-conditioned policies $\pi(a|s,z)$ using guided intrinsic motivation to explore and collect diverse experiences.
\end{itemize}

In the following sections, we explain how these components can be integrated into existing exploration methods to guide exploration towards preferred regions. 

\subsubsection{State Marginal Matching}
We base the exploration phase of our work on SMM (State Marginal Matching \cite{smm2019}), although our approach is not limited to this method. SMM aims to learn a state marginal distribution $\log \rho_{\pi_z}(s)$ to match a given target distribution $p^*(s)$ by minimising their Kullback-Leibler (KL) divergence. Additionally, to explore more efficiently, \citet{smm2019} proposed to learn latent-conditioned policies $\pi(a|s,z)$ by adding the diversity objective from \citet{Eysenbach2018}. Thus, the reward function is defined as:
\begin{eqnarray}
\label{eq:r_smm}
r_z(s)=r_z^{\text{exploration}}(s) + r_z^{\text{diversity}}(s)
\end{eqnarray}
where :
\begin{eqnarray}
\label{eq:r_ex}
r_z^{\text{exploration}}(s) = \underbrace{\log p^*(s)}_{(a)} -  \underbrace{\log \rho_{\pi_z}(s)}_{(b)} \\ r_z^{\text{diversity}}(s) =  \underbrace{\log q_\phi(z|s)}_{(c)} -  \underbrace{\log(p(z))}_{(d)}
\end{eqnarray}

Intuitively, according to \citet{smm2019}, the above equations imply that the agent should go to states (a) with high probability under the target state distribution, (b) where this agent has not been before, and (c) where this skill is clearly distinguishable from other skills. The last term (d) encourages exploration in the space of mixture components $z$.

\subsubsection{Adding reward from preferences}
\label{subsec: smm+r}

In order to direct the exploration towards preferred regions, we use $\hat{r}_{\psi}$  as the target distribution $p^*(s)$ in Equation \eqref{eq:r_ex} to motivate the agent to explore regions with high preference-based rewards. Therefore Equation \eqref{eq:r_ex} can be rewritten as: 
\begin{eqnarray}
\label{eq:r_ex_pref}
r_z^{\text{exploration}}(s,a) = \hat{r}_{\psi}(s)  - \log \rho_{\pi_z}(s) 
\end{eqnarray}

\subsubsection{Adding preferred regions and skills discovery}
\label{subsec: smm+sdpr}
Following the definition of preferred region and skill discovery described in Section \ref{sec:VQSD}, we define a potential preferred region $\hat S$ with  $\hat{r}_{\psi}$ according to (Equation \eqref{eq:pref_region}) on states collected online, and use it to train a discriminator $q_\phi$ presented in Section \ref{sec:VQSD}. The discriminator $q_\phi$, encourages each skill to explore distinct regions related to the potential preferred region. In other words, we incentivize the agent to learn diverse skills within the preferred region. Therefore, we define the diversity reward as:
\begin{eqnarray}
\label{eq:r_div_pref}
r_z^{\text{diversity}} = \log q_\phi(\hat s|z), \text{ with } \hat s \in \hat S
\end{eqnarray}

\subsubsection{Overall objective}
By combining each of the different reward components mentioned, the overall reward function to enable exploration towards preferred regions is given by: 
\begin{eqnarray}
\label{eq:rewskill}
r_z(s,a) = \underbrace{ \hat{r}_{\psi}(s)}_{(a)} -   \underbrace{\log \rho_{\pi_z}(s)}_{(b)}  +  \underbrace{\log q_\phi(\hat s|z)}_{(c)}
\end{eqnarray}

Intuitively, Equation \eqref{eq:rewskill} implies that the agent should go to (a) states with high preference rewards (b) states where the agent has not been before, and (c) to distinct regions within potential preferred regions. Our overall guided exploration method is described in Algorithm \ref{alg:guided_exploration}.

\subsubsection{Learning skills}
By following the previous objective in Equation \eqref{eq:rewskill} we explore the preferred region and train a discriminator $q_\phi$ to assign diverse regions of the preferred region to skills. We then use the discriminator $q_\phi$ to learn skills in the skill learning phase, as described in Section \ref{sec:SLD}.

\subsection{Preferred Latent Representation}
Despite being able to restrict diversity to specific regions of the environment, skills discovered in the state space might not appear diverse from a human point of view. The state space in MuJoCo \cite{Todorov2012} environments, for example, is a concatenation of joint positions and velocities. Discovering skills in this space often results in static positions. Even though easily distinguishable by the discriminator, they may seem similar to the human eye. Hence, as recommended by \citet{Eysenbach2018}, we examine using prior knowledge to identify discernably diverse skills.

 This prior can be represented as any function of the state space and used as a prior to condition the discriminator. In this case, the discriminator is defined as $q_\phi(f(s)|z)$ with $f(s)$ being the prior.

Although it can be useful to encourage the learning of specific types of skills by specifying a prior, relying on specifically designed priors may be limiting. Thus, we present an alternative to manually specifying this prior to learn skills that are more discernably diverse to human eyes. Specifically, we simply use the representation in the last hidden layer of the reward model $\hat{r}_{\psi}$ learnt from human preferences as the prior. 
The intuition is that the last hidden layer of the neural network that models the internal reward function of a human, should learn a latent state representation that captures features that matter for human preferences. We refer to this as the preferred latent representation.

Formally we can write $\hat{r}_{\psi}(s)$ as:
\begin{eqnarray}
\label{eq:latent_pref}
\hat{r}_{\psi}(s)=h_\psi(f_\psi(s))
\end{eqnarray}
where $f_\psi$, represents all layers of the reward model all layers except the output layer and $h_\psi$ is the output layer of the neural network. Hence, we define the discriminator in Equation \eqref{eq:r_div_pref} as  $q_\phi(f_\psi(\hat s)|z)$.

As depicted later in the experiments, this general approach for specifying priors achieves discernably diverse behaviors, while obviating the need for any additional training.

%%%%%%%%%%%%%%%%%%%%%%%%%%%%%%%%%%%%%%%%%%%%%%%%%%%%%%%%%%%%%%%%%%%%%%%%

\begin{algorithm}
\textbf{Initialize} $\mathcal B$ , $\pi_z, r_\psi, q_\phi$ \;
\ForEach{timestep}{
    Sample $z \sim p(z)$ \;
    // Collect data \;
    \For{each timestep $t$}{
        Sample action $a_t \sim \pi_\theta(a_t|s_t,z)$\;
        Step environment $s_{t+1} \sim p(s_{t+1} | s_t, a_t)$ \;
        Set reward $r_z(s)$ as in \eqref{eq:rewskill}\;%$ = r_\psi(s) - \log \rho_{\pi_z}(s)  + \log q_\phi(s_{t+1}|z) $\;
        Update policy $(\theta)$ to maximise $r_z$ with SAC \cite{pmlr-v80-haarnoja18b}\;
        Store transitions $\mathcal B \leftarrow \mathcal{B}\cup \{(s_t,a_t,s_{t+1}, r_z)\}$ \;
    }
    \If { it's time to update the preference} {   
        // Query instructor \;
        \ForEach{query to instructor}{
            Sample $(\sigma^0,\sigma^1) \sim \mathcal B $ \;
            Collect preference from instructor $y=\sigma^0 \succ \sigma^1$ \;
            Store transitions $\mathcal D \leftarrow \mathcal{D}\cup \{(\sigma^0,\sigma^1, y)\}$
                }  
        // Update reward model \;
        \ForEach{each gradient step}{
            Sample minibatch $(\sigma^0,\sigma^1, y)^{\mathcal D}_{j=1}\sim \mathcal D $ \;
            Optimize $\mathcal L^{\text{reward}} $ in \eqref{eq:BradleyTerryLoss} with respect to $\psi$
            }
        // Estimate the preferred region  \;
        $\hat S = \{\forall s \in \mathcal{B} | \hat {r}_\psi(s) \geq \beta \}$\; 
        
        // Skill Discovery phase \;
        \ForEach{query to instructor}{
        Sample minibatch $ (s)^{\hat S}_{j=1} \sim \hat S $ \;
        Optimize $\mathcal L^{\text{VQ-VAE}} $ in \eqref{eq:vq-vaeloss} with respect to $\phi$           
        }
        
    }
}
\caption{Guided exploration with preferences}
\label{alg:guided_exploration}
\end{algorithm}

\section{Experiments}

In this section, we examine our proposed method to control diversity with preference and to guide the agent's exploration towards preferred regions.
We first demonstrate our approach on a 2D navigation environment, following which we also show the performance of our method in higher dimensional environments such as MuJoCo in Section \ref{sec:preflatentExp} and \ref{sec:beta}. The 2D environment consists of a two-dimensional room enclosed by walls that restrain the agent. The agent begins each episode in the middle of the room, until episode termination, which occurs after $100$ steps. The agent has only access to its horizontal and vertical coordinates (X,Y). It can deterministically change its direction and amplitude of steps to freely move in the environment. Both state space and action space are continuous.
Following previous work on preference-based RL, we simulate human preference with an oracle 'true' reward function. The true reward function is designed to be a gaussian distribution, centered around a goal position, and the reward is computed as the negative distance to the goal.

\subsection{Results in 2D navigation}
\label{sec: cdpexp}
In the 2D navigation environment, we intend to demonstrate that a preferred region can be used to define a relevant area of interest. In the interest of studying the effectiveness of preferred regions for discovering skills, in this section, we assume an ideal state coverage and an oracle reward function. We show both EDL and CDP results to demonstrate the full impact of the preferred region.

\begin{figure}[ht]
    \centering
    \begin{subfigure}[b]{0.405\textwidth}
        \centering
        \includegraphics[width=\textwidth]{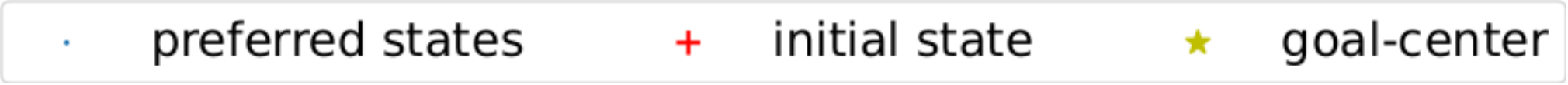} 
    \end{subfigure}
    \begin{subfigure}[b]{0.2\textwidth}
        \centering
        \includegraphics[width=\textwidth]{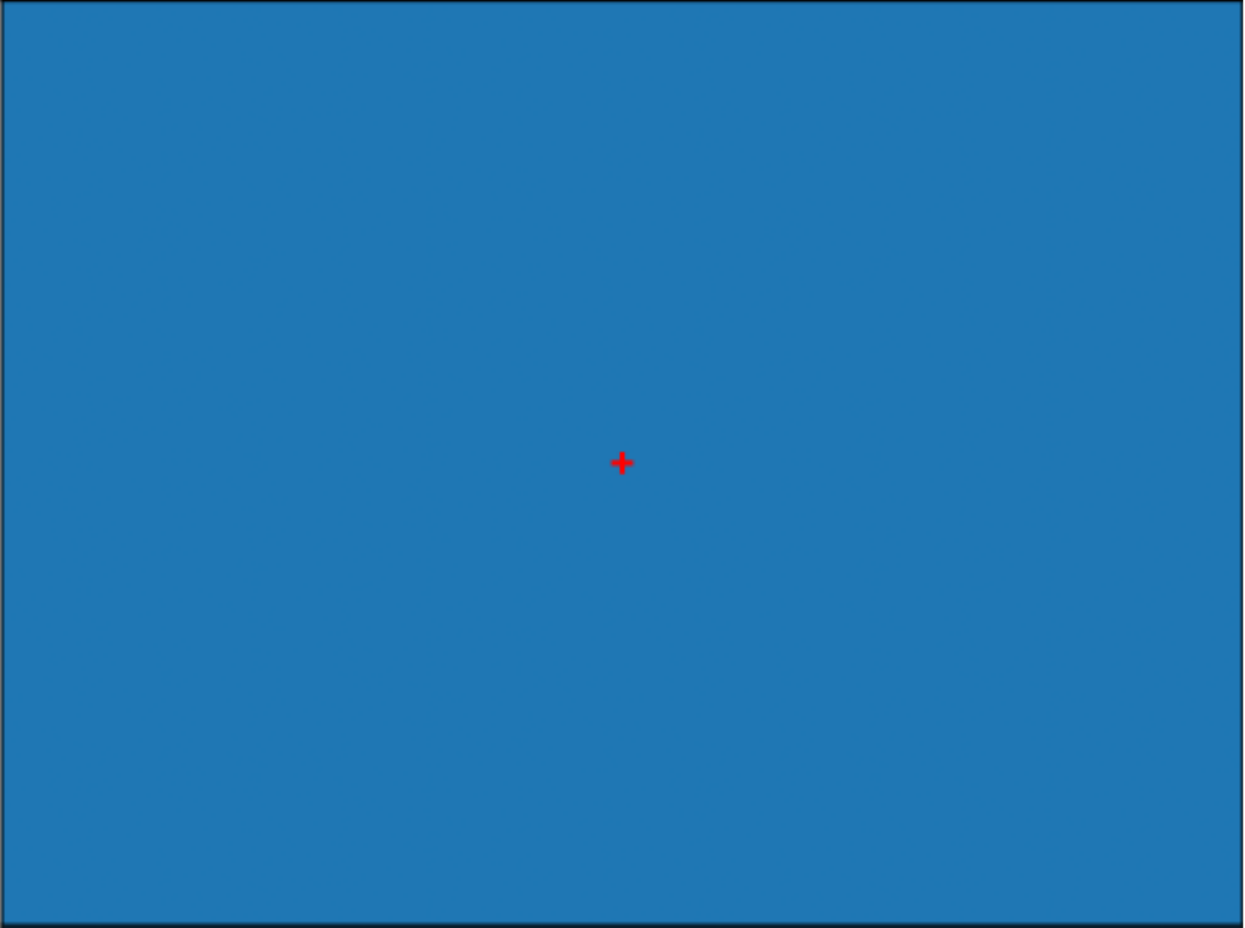}
        \caption{Full state space}
        \Description{The full state space is covered in bleu to represent that it will be used for everything}
        \label{fig: ex-edl}
    \end{subfigure}
    \begin{subfigure}[b]{0.2\textwidth}
        \centering
        \includegraphics[width=\textwidth]{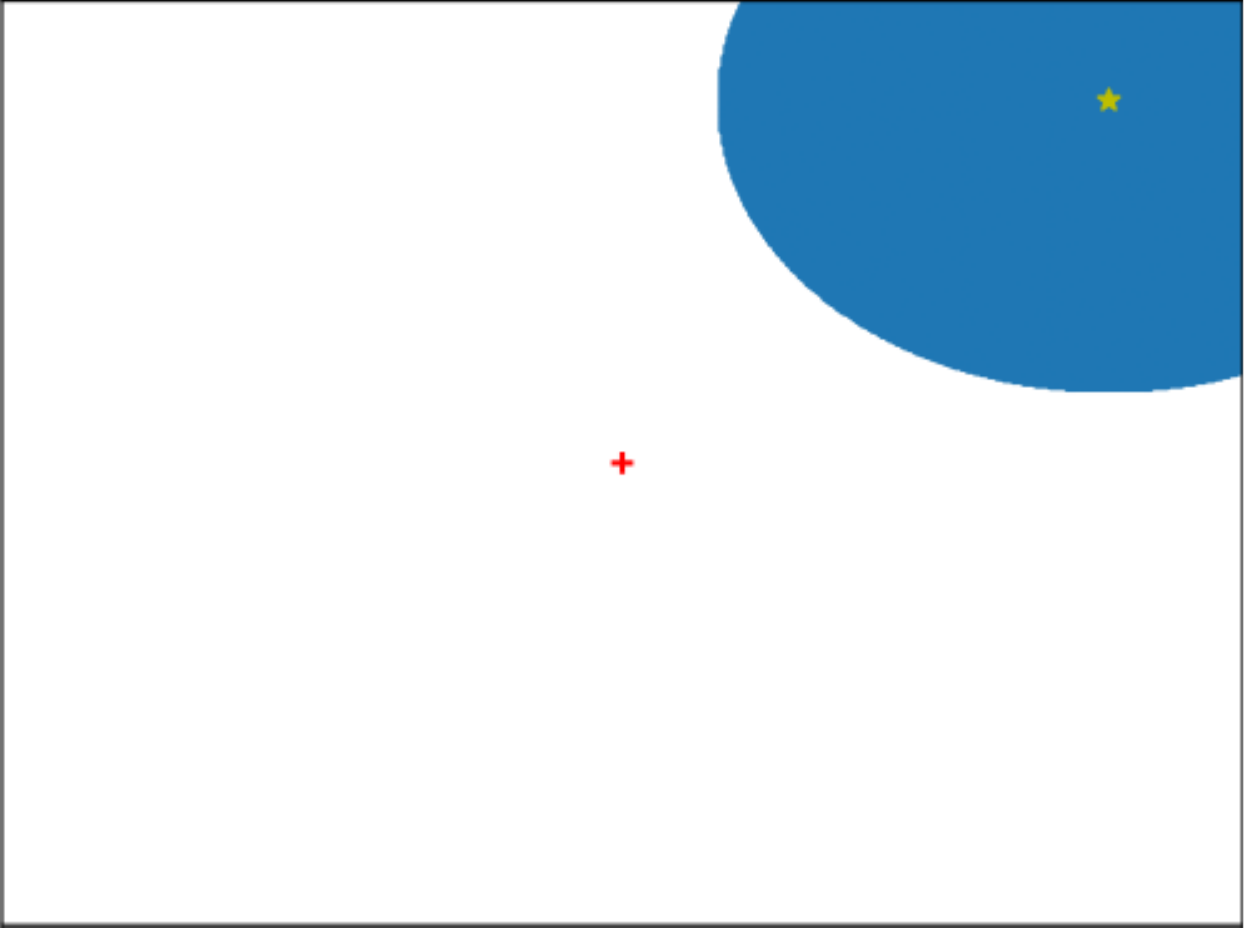} 
        \caption{CDP-preferred region}
        \Description{The preferred region of the state space obtained by our methods is covered in blue and located in the top right corner}
        \label{fig: ex-cdp}
     \end{subfigure}
     \begin{subfigure}[b]{0.405\textwidth}
        \centering
        \includegraphics[width=\textwidth]{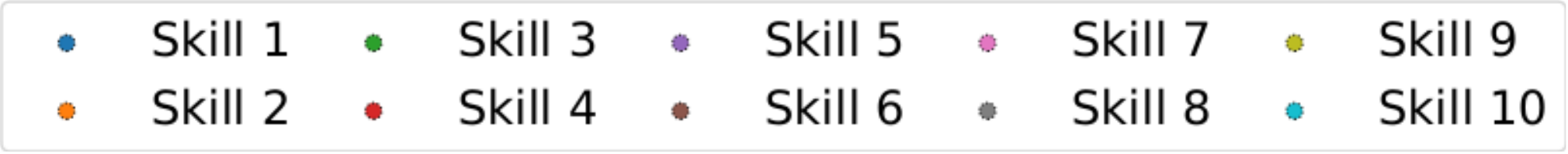} 

    \end{subfigure}
     \begin{subfigure}[b]{0.2\textwidth}
         \centering
         \includegraphics[width=\textwidth]{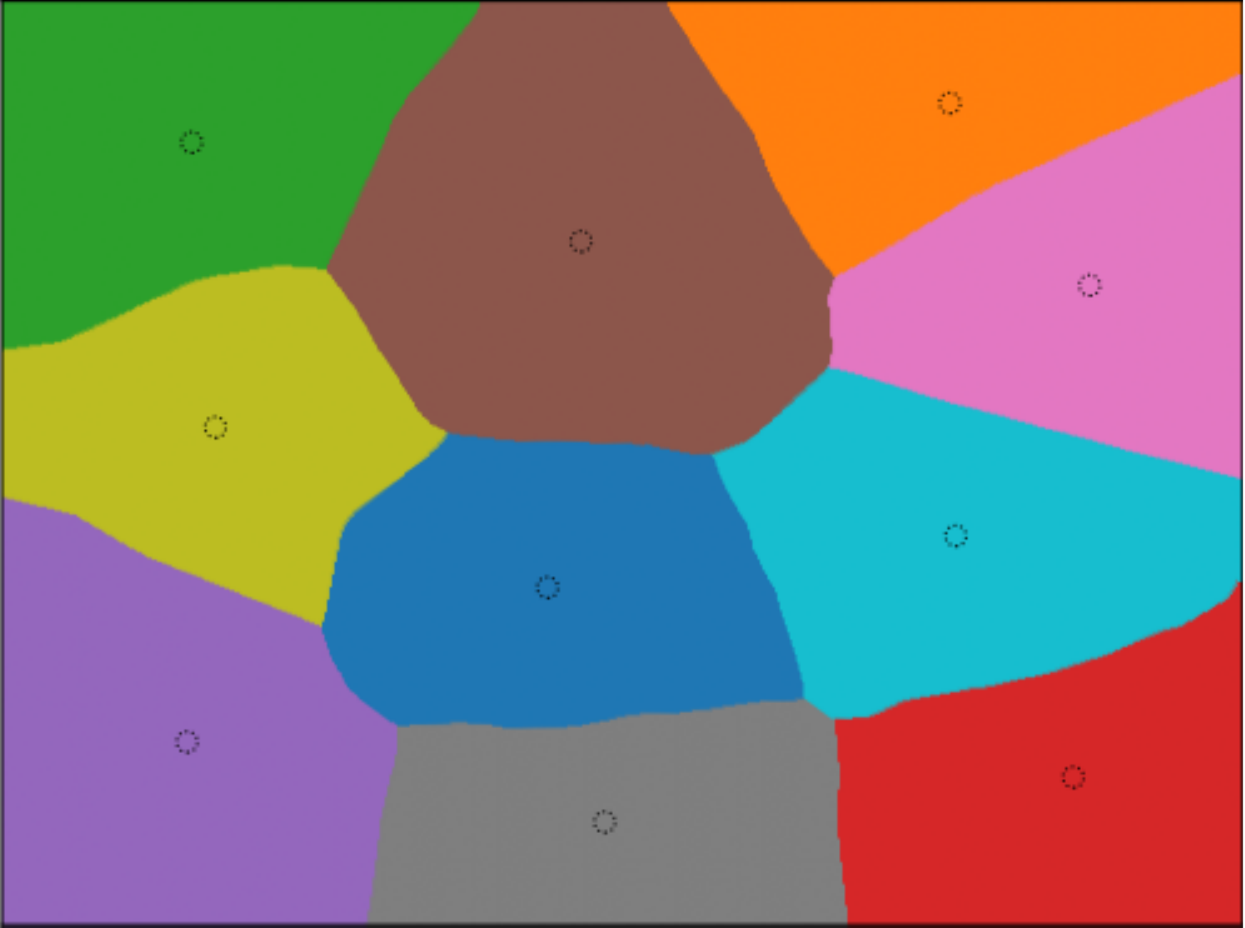} 
         \caption{EDL-assigned regions}
         \Description{The full state space is divided in 10 regions uniformly across the state space.}
         \label{fig: skillR-edl}
     \end{subfigure}
    \begin{subfigure}[b]{0.2\textwidth}
         \centering
         \includegraphics[width=\textwidth]{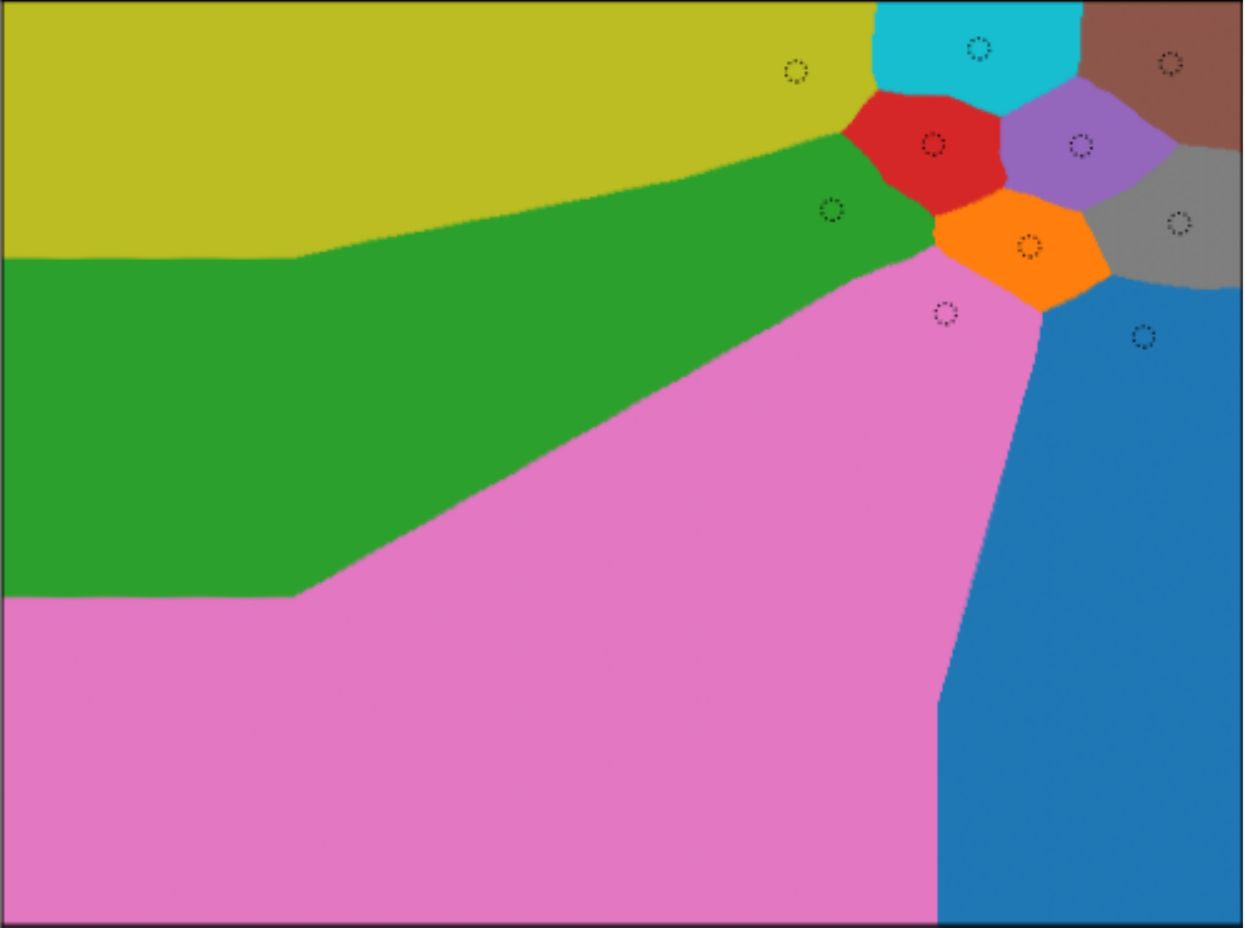} 
         \caption{CDP-assigned regions}
        \Description{The full state space is divided in 10 regions uniformly across the top right corner.}
         \label{fig: skillR-cdp}
     \end{subfigure}
    \begin{subfigure}[b]{0.405\textwidth}
        \centering
        \includegraphics[width=\textwidth]{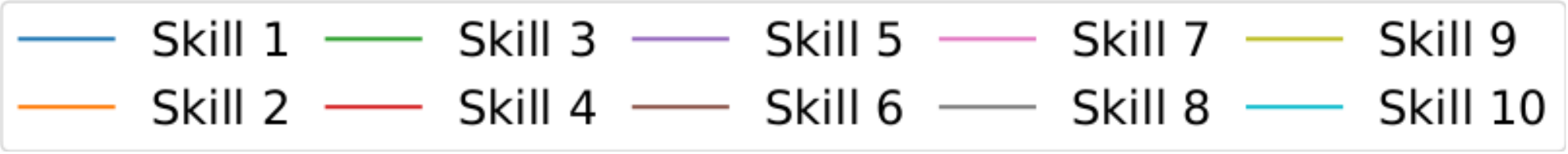} 
    \end{subfigure}
     \begin{subfigure}[b]{0.2\textwidth}
         \centering
         \includegraphics[width=\textwidth]{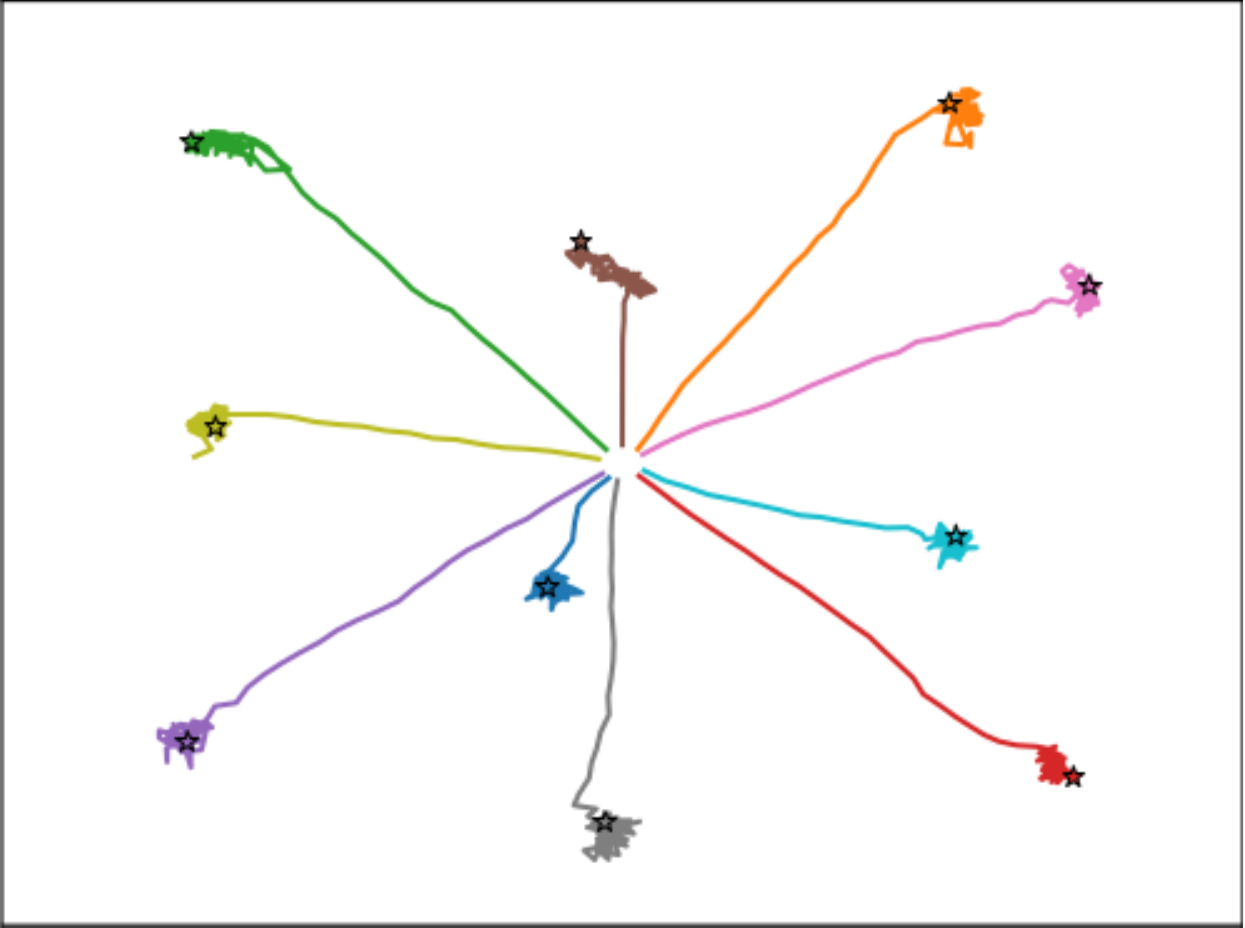} 
         \caption{EDL-learned skills}
         \Description{The figure shows that the skills learned are moving in all directions.}
         \label{fig: skill-edl}
     \end{subfigure}
     \begin{subfigure}[b]{0.2\textwidth}
         \centering
         \includegraphics[width=\textwidth]{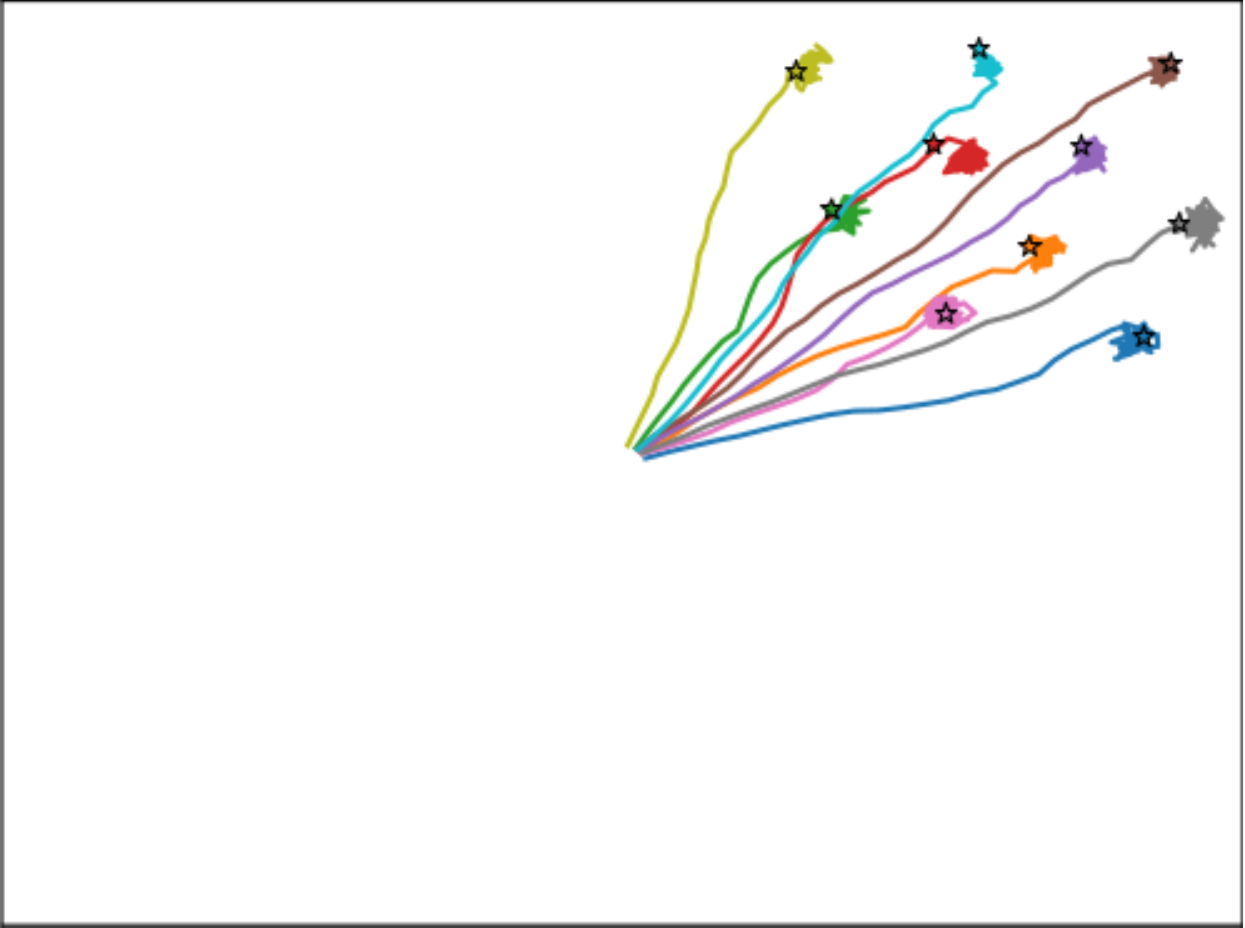} 
         \caption{CDP-learned skills}
         \Description{The figure shows that the skills learned are moving toward the top right corner.}
         \label{fig: skill-cdp}
     \end{subfigure}
        \caption{(a), (c) and (e) are respectively the full state space, the regions assigned to each skill by the discriminator trained on the full state space and the skills learned with this discriminator. (b), (d) and (f) are respectively the preferred regions of the state space obtained by our method, the regions assigned to each skill by the discriminator trained on the preferred region and the skills learned with this discriminator.}
        \Description{(a), (c) and (e) are respectively the full state space, the regions assigned to each skill by the discriminator trained on the full state space and the skills learned with this discriminator. (b), (d) and (f) are respectively the preferred regions of the state space obtained by our method, the regions assigned to each skill by the discriminator trained on the preferred region and the skills learned with this discriminator.}
     \label{fig: tc}
\end{figure}

By applying the definition of the preferred region described in Section \ref{sec:PR} to the assumed state coverage in Figure \ref{fig: ex-edl}, we identify the preferred region in the top right corner, as indicated in Figure \ref{fig: ex-cdp}.
We then discover and learn skills in those proposed regions. As illustrated in Figure \ref{fig: skillR-edl} and \ref{fig: skill-edl}, EDL discovers and learns skills uniformly across the environment. In our case (CDP), the discriminator concentrates all skills' assigned regions in the top right corner, as shown in Figure \ref{fig: skillR-cdp}. Additionally, in Figure \ref{fig: skillR-cdp}, centroids (the most likely state under the discriminator for each skill) are located in the top corner, resulting in skills moving to the top right corner as illustrated in Figure \ref{fig: skill-cdp}. 

\subsection{Exploration of the Preferred Region}
\label{sec: gepexp}
This section aims to demonstrate that the modifications we made to the SMM method in Section \ref{sec:gep} have significant advantages with regards to exploring the preferred region. We place ourselves in more realistic settings where we don’t have full state coverage, or have access to the oracle reward. We compare our method with SMM as described in EDL and SMM+prior that uses the same prior as us. The prior is a reward function learned from preference, and used as described in Section \ref{subsec: smm+r}. 

\begin{figure}[ht]
    \centering
    \begin{subfigure}[b]{0.3\textwidth}
         \centering
        \includegraphics[width=\textwidth]{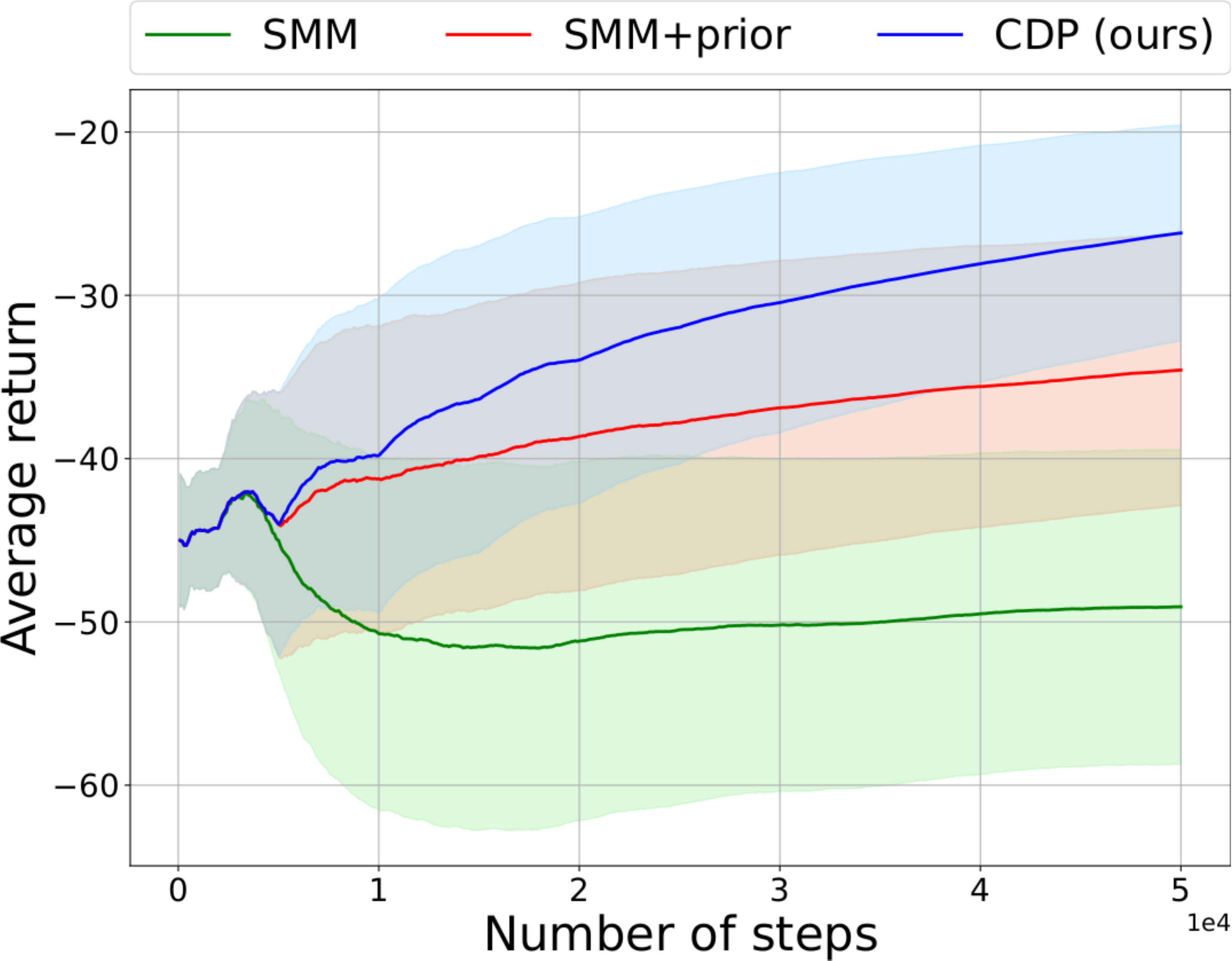} 
    \end{subfigure}
    \caption{Average return achieved by each method.} 
     \Description{The figure compares the average return achieved by each method across training time}
    \label{fig:rew}
\end{figure}

As illustrated in Figure \ref{fig:rew}, we compared each method in terms of their average returns as per the target reward function. Intuitively, exploring more of the preferred region should result in a higher return. Results in Figure \ref{fig:rew} suggest that our proposed method visits more states with higher rewards than the other methods, which implies that it explores the preferred region more efficiently.

\begin{figure}[ht]
     \centering
      \begin{subfigure}[b]{0.33\textwidth}
         \centering
         \includegraphics[width=\textwidth]{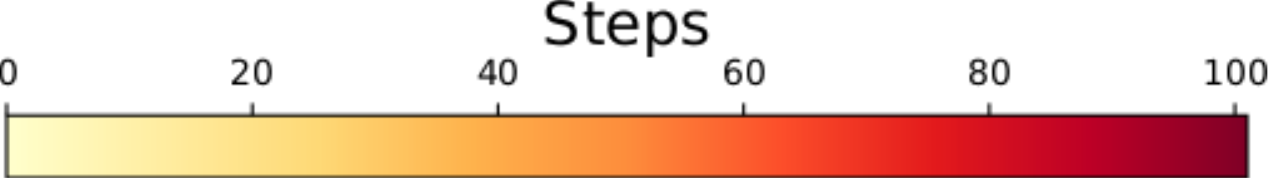} 
     \end{subfigure}
     \begin{subfigure}[b]{0.15\textwidth}
         \centering
         \includegraphics[width=\textwidth]{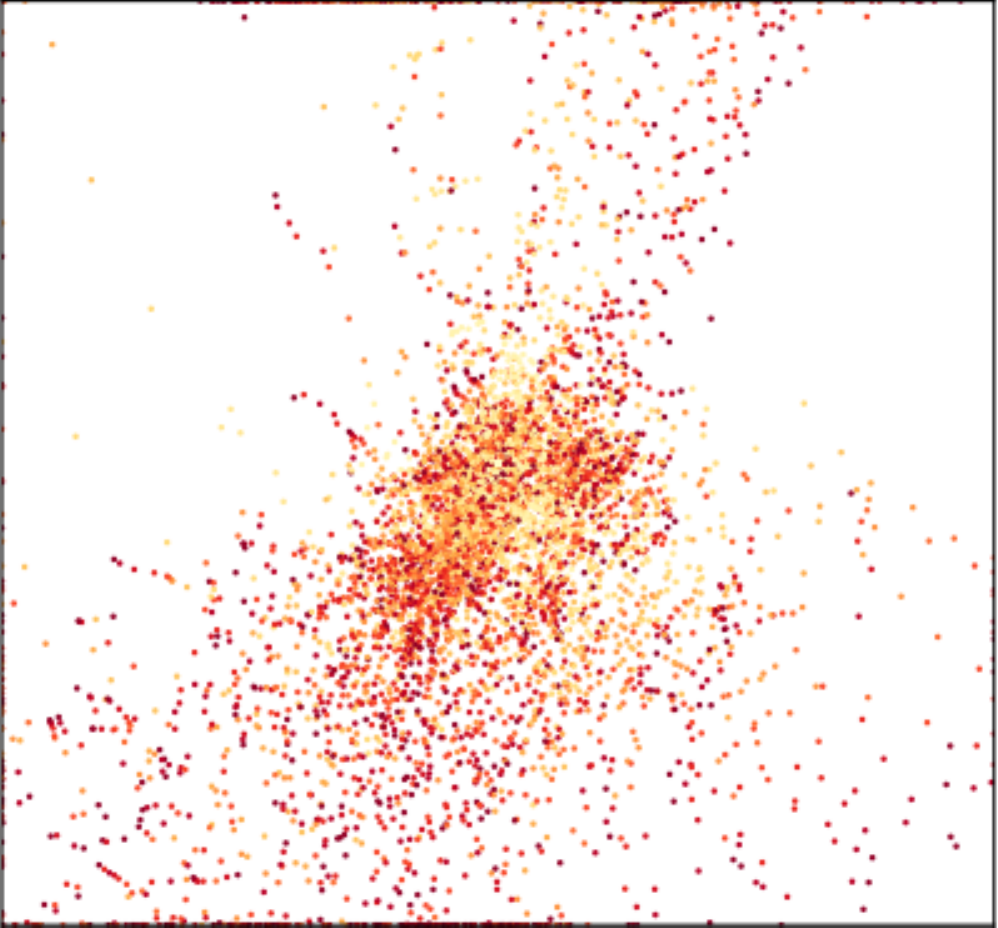} 
         \label{fig:ex1}
         \caption{SMM}
         \Description{The figure shows the states visited during the exploration. Most of the states visited are in the center}
     \end{subfigure}
     \begin{subfigure}[b]{0.15\textwidth}
         \centering
         \includegraphics[width=\textwidth]{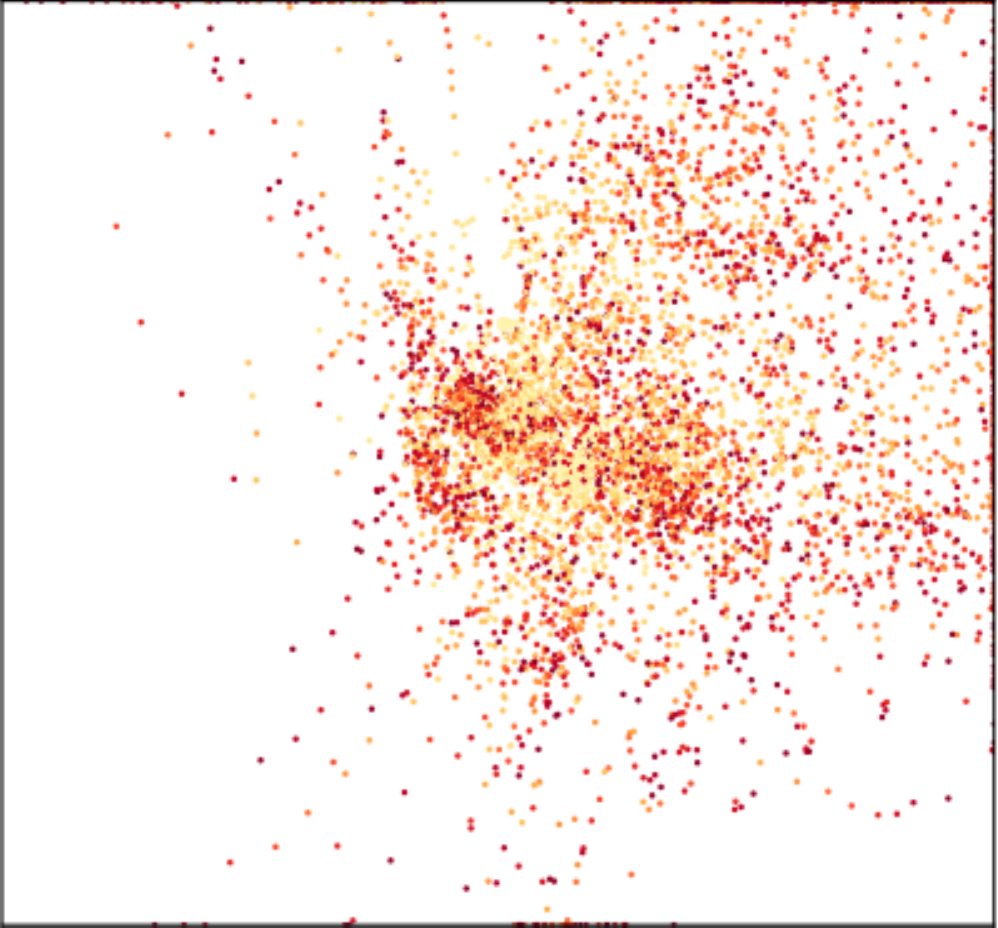} 
         \label{fig:ex2}
         \caption{SMM+prior}
         \Description{The figure shows the states visited during the exploration. Most of the states visited are in the center and in the top right corner}
     \end{subfigure}
     \begin{subfigure}[b]{0.15\textwidth}
         \centering
         \includegraphics[width=\textwidth]{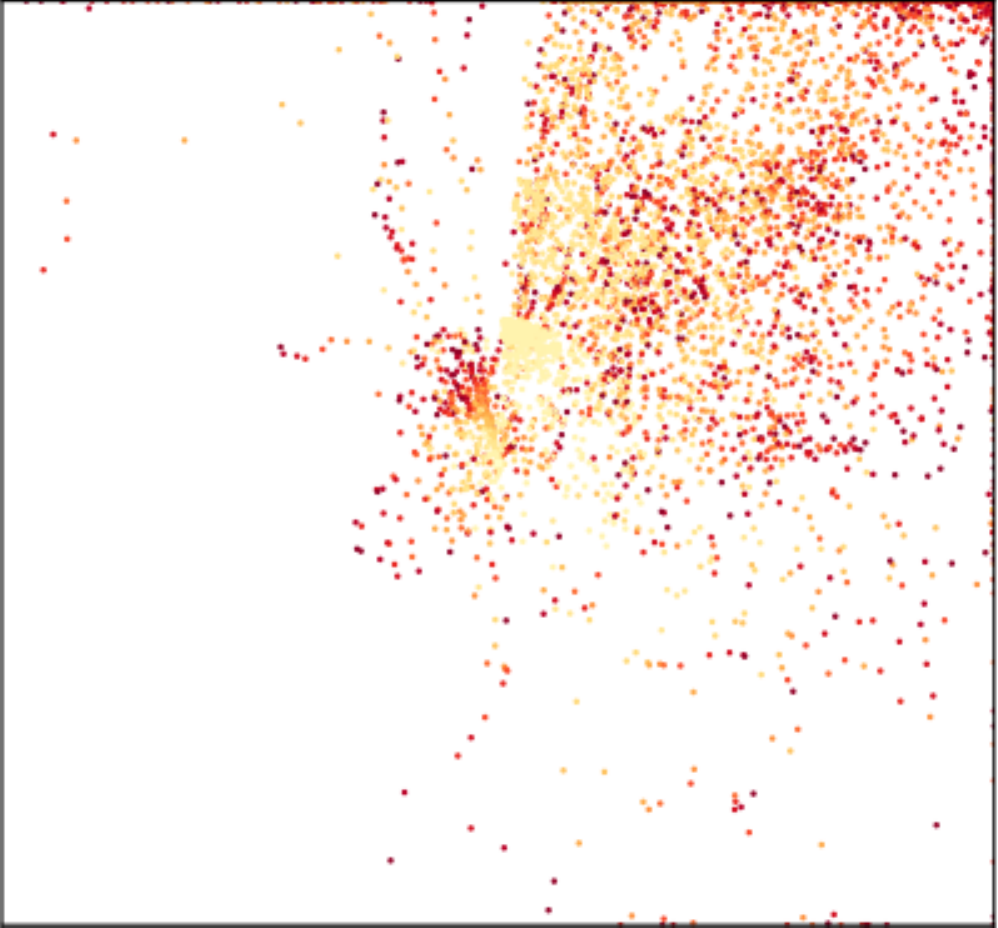}
         \label{fig:ex3}
         \caption{CDP (ours)}
         \Description{}
     \end{subfigure}
        \caption{States visited by each of the methods, SMM (a), SMM+prior(b), ours(c).}
        \label{fig:exploration}
         \Description{The figure shows the states visited during the exploration. Most of the states visited are in the top right corner}
\end{figure}

From a qualitative perspective, Figure \ref{fig:exploration} depicts the states visited by each method and shows that our method visits more states in the top right corner (preferred region). Further, the presence of darker shades (indicating the later stages of interaction) in the top right corner indicates that the skills from our method tend to end near or in the preferred region. This can be explained by the discriminator incentivizing the agent to learn diverse skills within the preferred region. This is in contrast to the other methods in which the discriminator only encourages agents to acquire diverse skills, as indicated by the darker points in Figure \ref{fig:exploration}a and \ref{fig:exploration}b being relatively more evenly distributed in different state regions, and not particularly within the preferred region.

The comparisons with the first method are unfair since they do not have access to any information about the preferred region. In spite of this, we still feel that the comparison is relevant to emphasize the choice to use human preference to control diversity.

\subsection{Results using Preferred Latent Representations}
\label{sec:preflatentExp}

In this section, we demonstrate that preferred latent representations facilitate the acquisition of appropriate skills in a general manner, that are capable of scaling to larger state and action spaces. To this end, we performed experiments on a MuJoCo-based modified Half Cheetah agent, in which moving backwards was preferred. It was specified by using a version of the original half-cheetah reward function modified (by multiplying the original reward by -1) to encourage the agent to move backwards as far as possible along the horizontal axis. In other words, we aim to achieve diverse velocities corresponding to the desired behavior of moving backwards.

\begin{figure}[ht]
     \centering
     \begin{subfigure}[b]{0.4\textwidth}
         \centering
         \includegraphics[width=\textwidth]{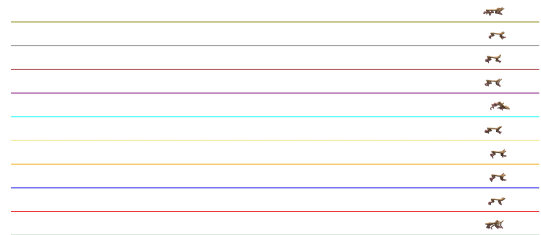} 
         \caption{Modified Half cheetah's skill learned using the state space to discover skills }
         \Description{The figure shows that the skills learned are not moving}
         \Description{}
         \label{fig: PLR-state}
     \end{subfigure}
     \begin{subfigure}[b]{0.4\textwidth}
         \centering
         \includegraphics[width=\textwidth]{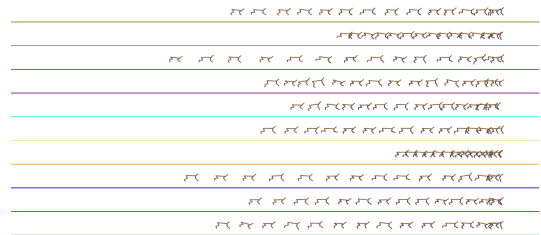} 
         \caption{Modified Half cheetah's skill using the preferred latent representation to discover skills}
         \Description{The figure shows that the skills learned are moving backward in different ways.}
         \label{fig: PLR-plr}
     \end{subfigure}
     \begin{subfigure}[b]{0.4\textwidth}
         \centering
         \includegraphics[width=\textwidth]{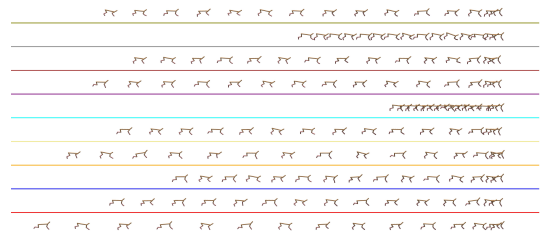} 
         \caption{Modified Half cheetah's skill using a manually specified prior over velocity to discover skills}
        \Description{The figure shows that the skills learned are moving backward at varying velocity.}
         \label{fig: PLR-speed}
     \end{subfigure}
    \begin{tikzpicture}
          \coordinate (A) at (-3,0);
          \coordinate (B) at ( 3,0);
          \draw[<-] (A) -- (B) node[midway,fill=white] {\emph{Direction of motion}};
    \end{tikzpicture}
        \caption{ Modified Half cheetah's skill learned using different representations of the state space to discover skills}
        
        \label{fig: PLR}
\end{figure}

As shown in Figure \ref{fig: PLR-state}, without additional prior knowledge about the state space, the agent does not learn any relevant skills. However, when using the preferred latent representation (Figure \ref{fig: PLR-plr}), the agent is able to learn diverse skills that go backward at varying speeds similar to skills learned while using a manually-specified prior (the agent's velocity) over velocity (Figure \ref{fig: PLR-speed}).

\begin{figure}[ht]
    \centering
    \begin{subfigure}[b]{0.25\textwidth}
        \centering
        \includegraphics[width=\textwidth]{../figures/cdp/exp1/legends/Skill.pdf} 
    \end{subfigure}
     \begin{subfigure}[b]{0.25\textwidth}
         \centering
         \includegraphics[width=\textwidth]{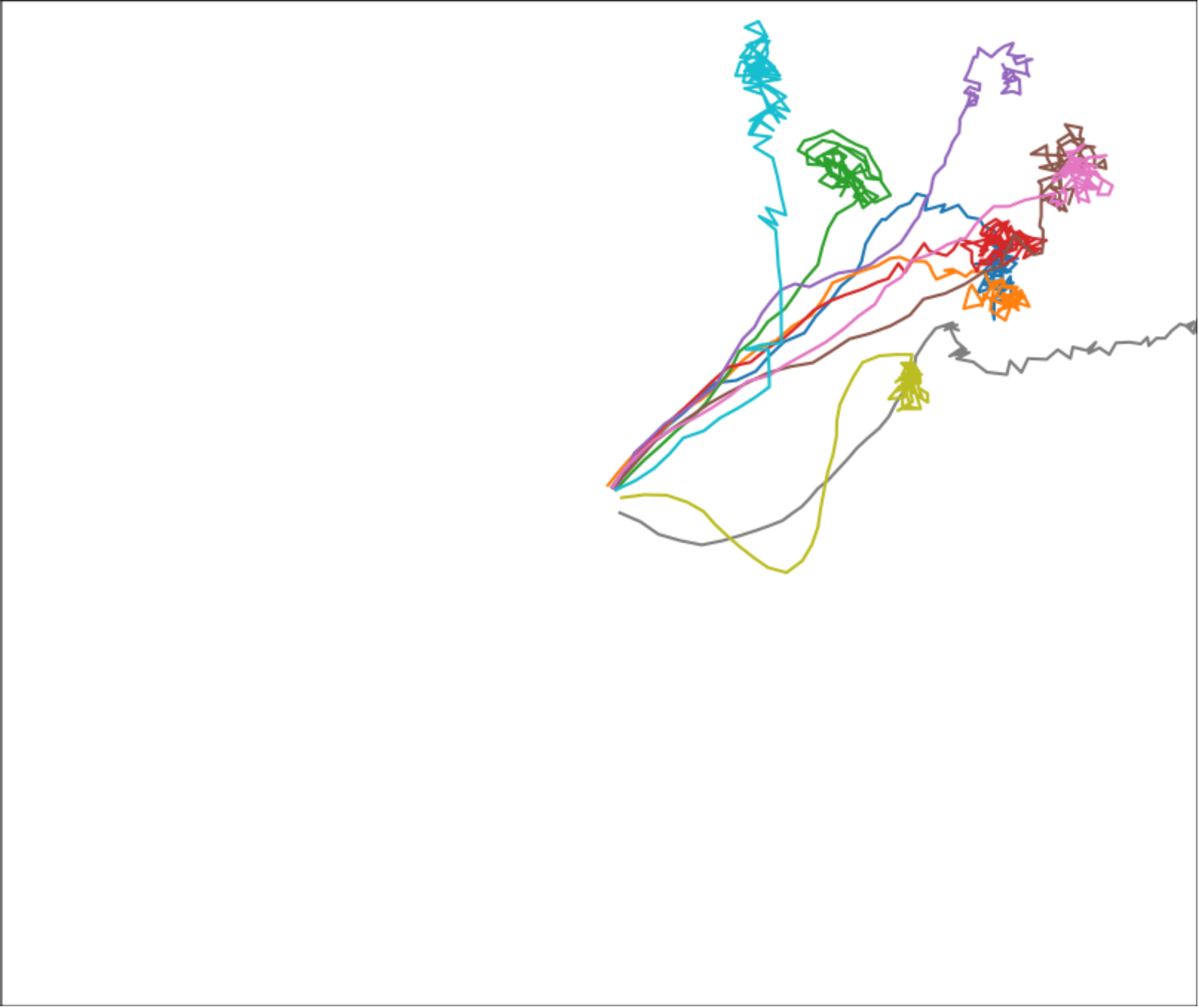} 
     \end{subfigure}
    \caption{Skill learned using the preferred latent representation as a prior to discover skills.}
    \Description{The figure shows that the skills learned are moving toward the top right corner.}
    \label{fig: PLR-nav}   
\end{figure}

Additionally, we repeat the 2D navigation experiments from Section \ref{sec: cdpexp}, but using preferred latent representations to learn diverse and desirable skills. As seen in Figure \ref{fig: PLR-nav} the agent learns skills comparable to those in Section \ref{sec: cdpexp}. We note that the trajectories in Figure \ref{fig: PLR-nav} are relatively more noisy when compared to those in Figure \ref{fig: skill-cdp}, probably due to the inherent noise associated with learning the preferred latent representations. However, the fact that preferred latent representations also enable the agent to learn the intended skills implies that they do indeed capture relevant features of the state space, be it in the navigation task, or the more complex backwards Half Cheetah environment.

\subsection{Effect of $\beta$}
\label{sec:beta}

\begin{figure}[ht]
     \centering
     \begin{subfigure}[b]{0.475\textwidth}
        \centering
        \includegraphics[width=\textwidth]{../figures/cdp/exp1/legends/SkillRegion.pdf} 
    \end{subfigure}
    \begin{subfigure}[b]{0.155\textwidth}
         \centering
         \includegraphics[width=\textwidth]{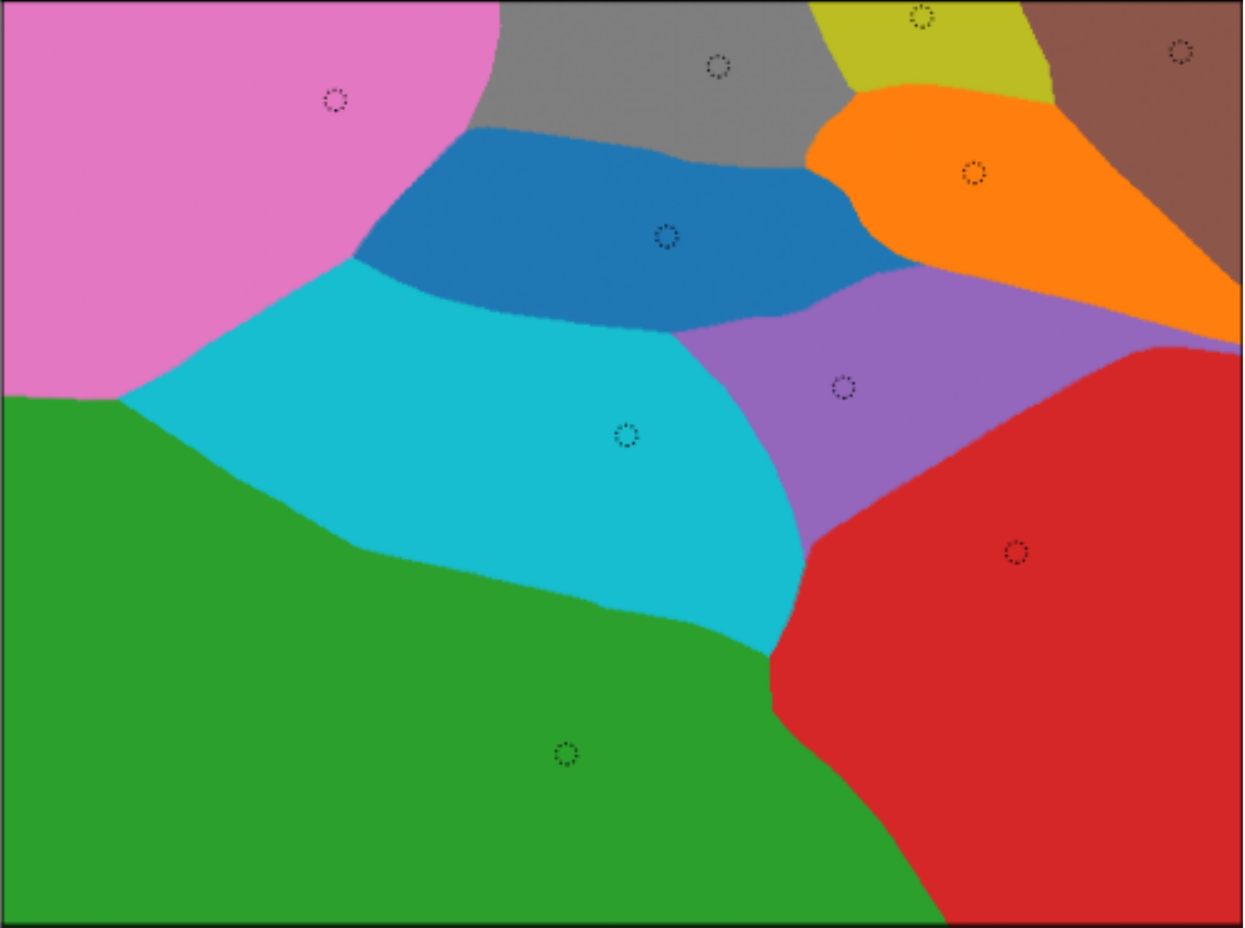} 
         \caption{$\beta$=0.1}

     \end{subfigure}
     \begin{subfigure}[b]{0.155\textwidth}
         \centering
         \includegraphics[width=\textwidth]{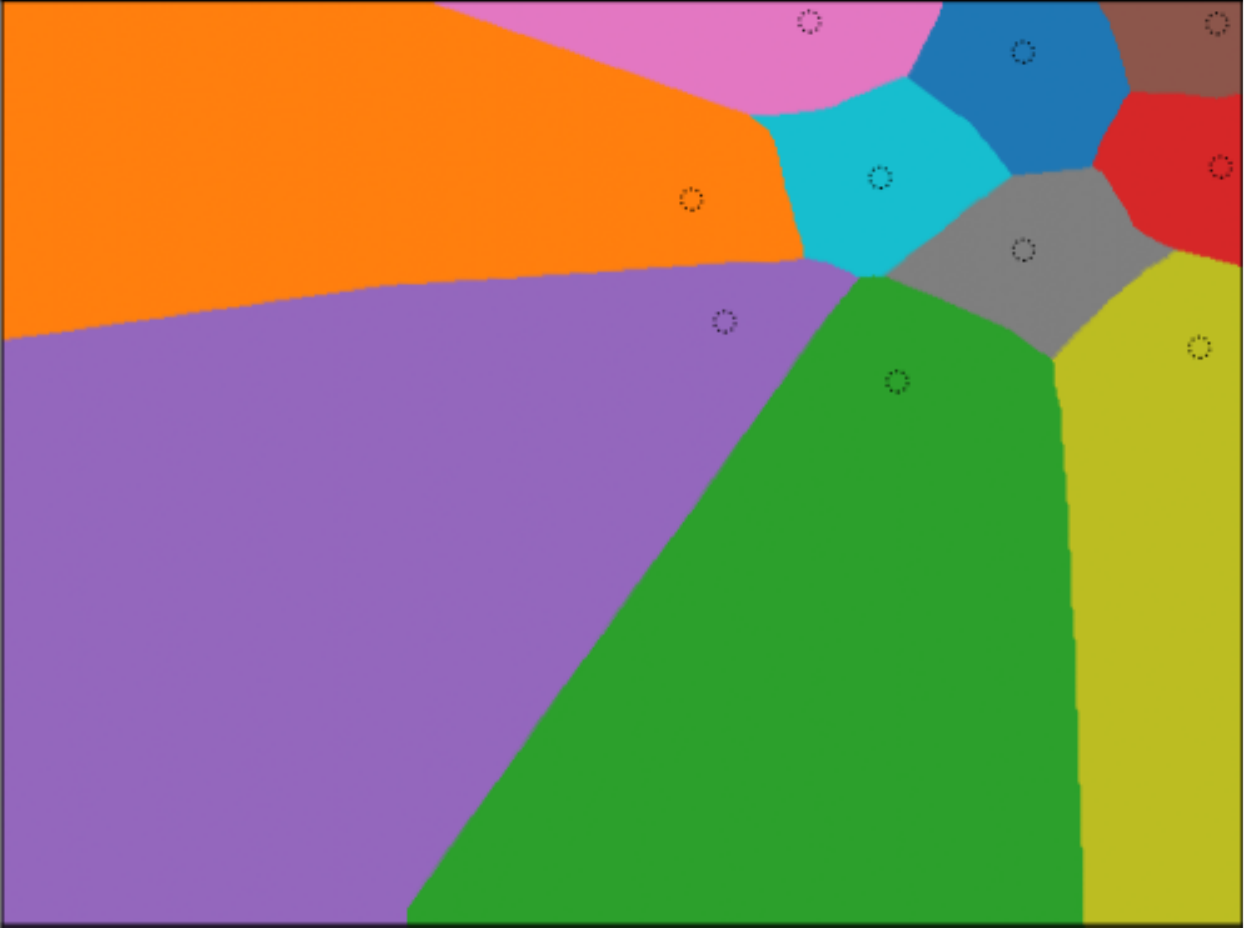} 
         \caption{$\beta$=0.5}
         \Description{The figure shows that the skills learned are moving backward at varying velocity over a short distance.}
     \end{subfigure}
     \begin{subfigure}[b]{0.155\textwidth}
         \centering
         \includegraphics[width=\textwidth]{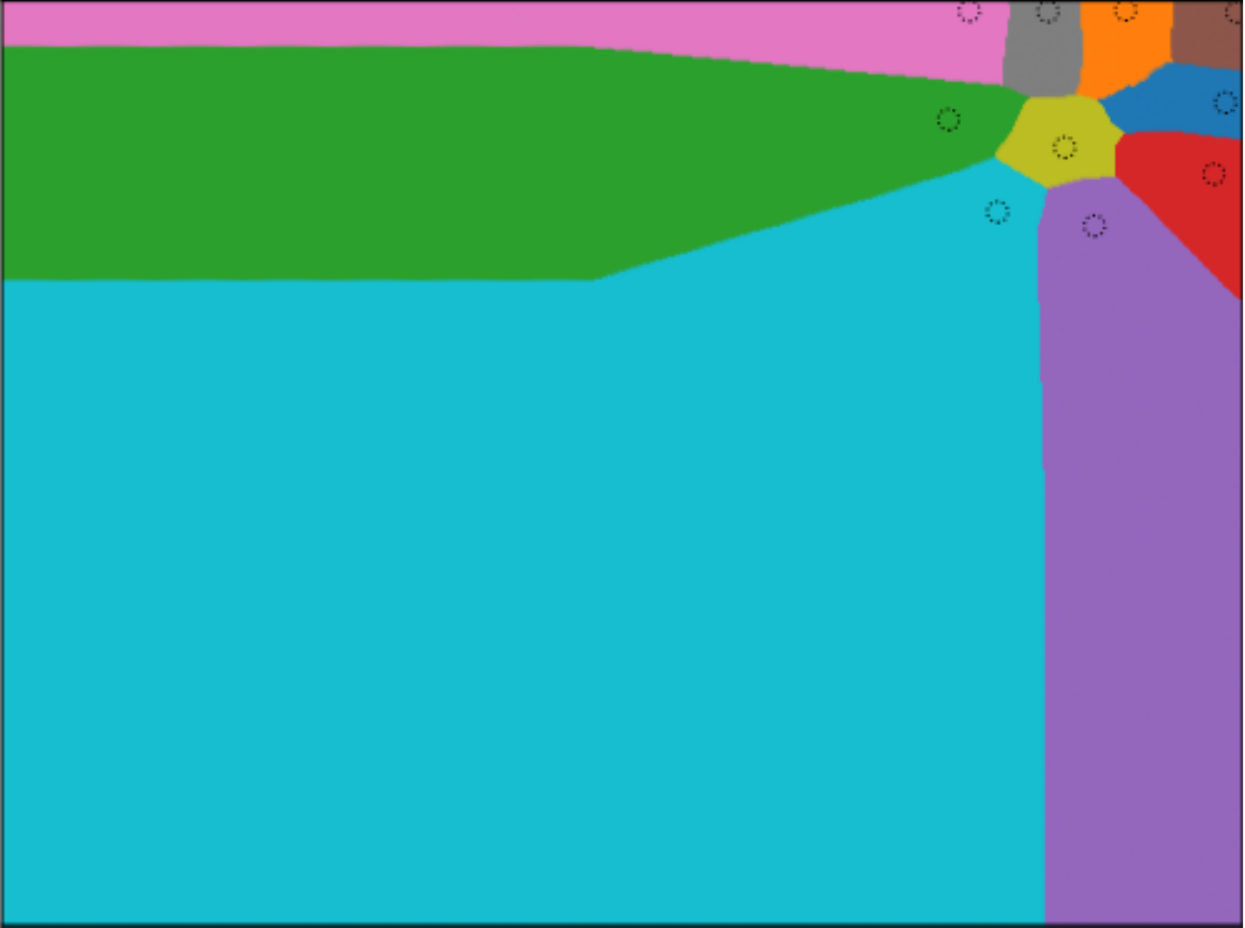} 
         \caption{$\beta$=0.9}
         \Description{}
     \end{subfigure}
        \caption{Regions assigned to each skill by the discriminator trained on preferred regions set by different values of $\beta$.}
        \Description{}
        \label{fig: beta1}
\end{figure}

Here, we examine the effect of varying $\beta$ (used in Equation \eqref{eq:pref_region}) on the resulting skills obtained. We show results for both the MuJoCo-based modified Half Cheetah and 2D navigation experiments. 
$\beta$ determines how much emphasis is placed on skill discovery centered around high rewards. This can be viewed as a parameter that controls how much we exploit the reward function to constrain skill discovery. In Figure \ref{fig: beta1}, we use the setting described in Section \ref{sec: cdpexp} to show that a low $\beta$ will produce skills that may be far away from the goal, while a high $\beta$ will learn skills around the goal. 

\begin{figure}[ht]
     \centering   
     \begin{subfigure}[b]{0.475\textwidth}
         \centering
         \includegraphics[width=\textwidth]{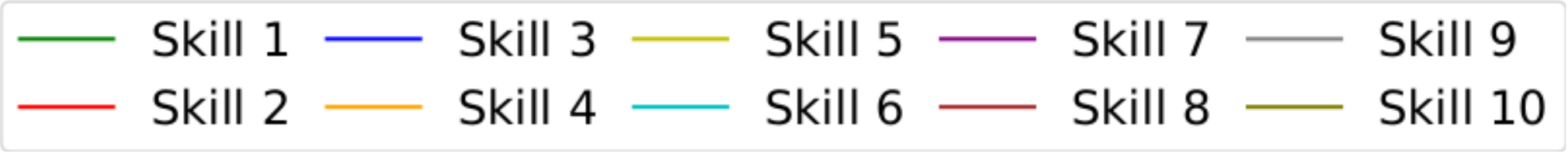} 
     \end{subfigure}

     \begin{subfigure}[b]{0.155\textwidth}
         \centering
         \includegraphics[width=\textwidth]{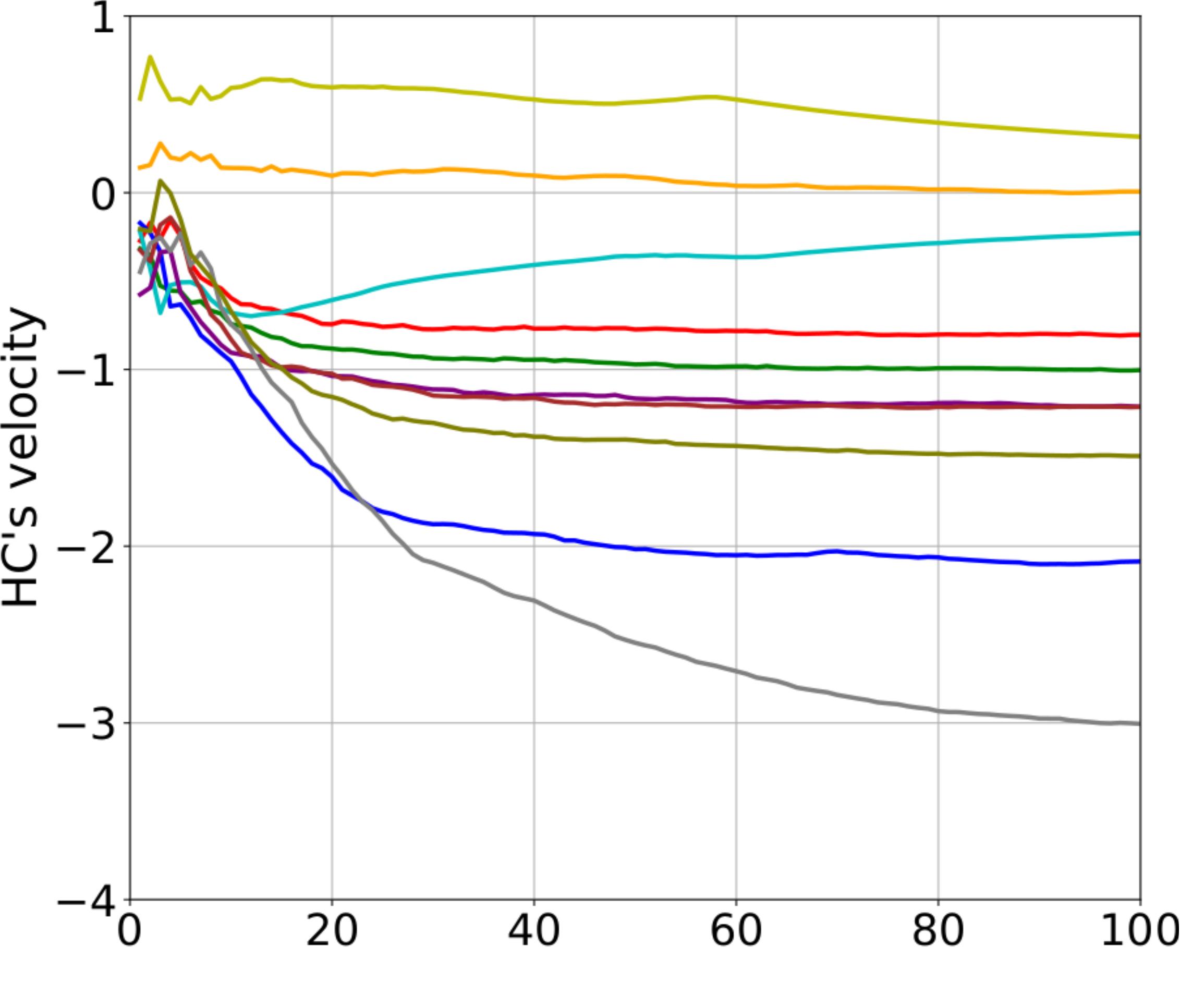} 
         \caption{$\beta$=0.3}
         \Description{The full state space is divided into 10 regions concentred in the top right of the state space.}
     \end{subfigure}
     \begin{subfigure}[b]{0.155\textwidth}
         \centering
         \includegraphics[width=\textwidth]{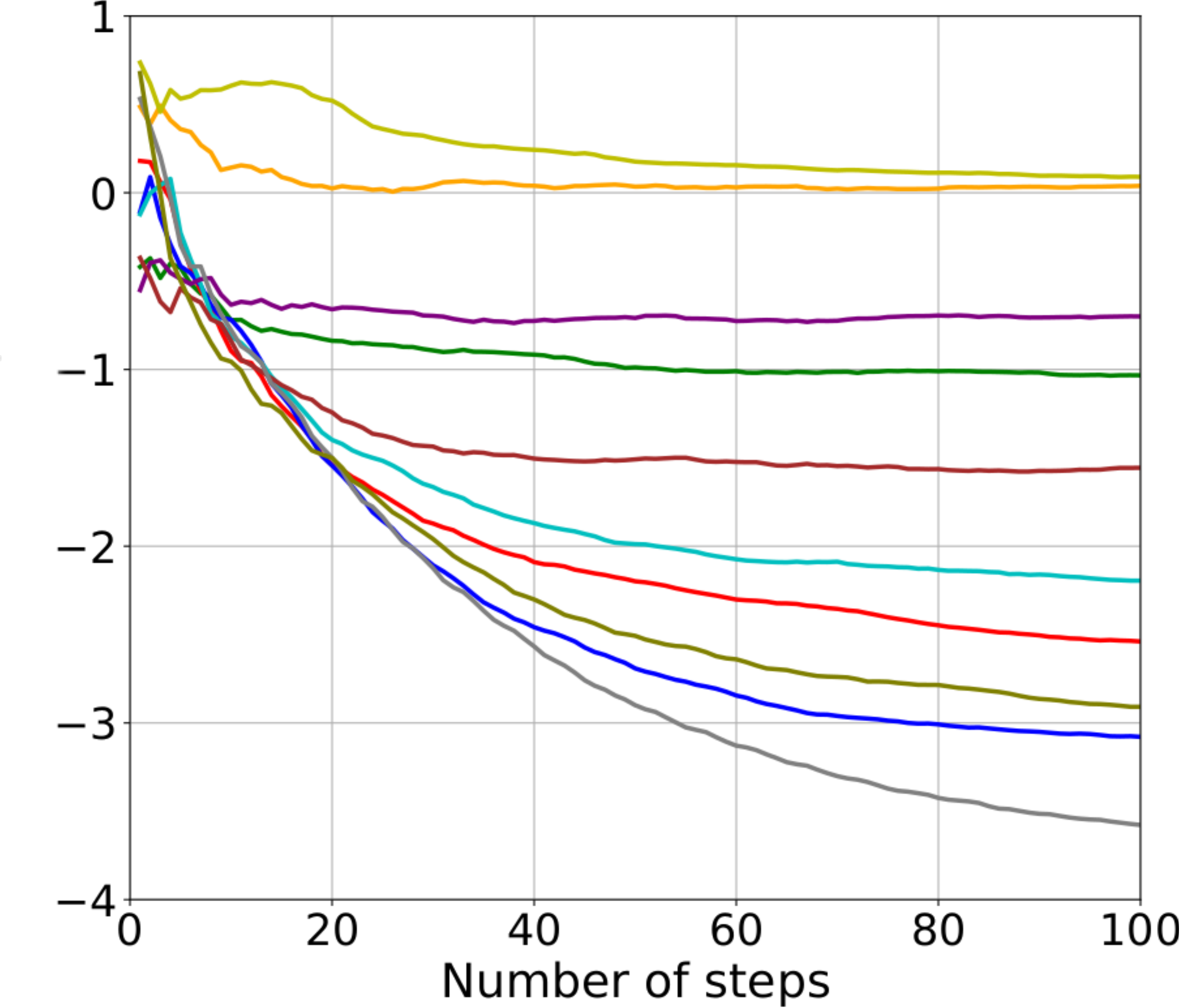} 
         \caption{$\beta$=0.5}
        \Description{The full state space is divided into 10 regions uniformly across the top right corner.}
     \end{subfigure}
     \begin{subfigure}[b]{0.155\textwidth}
         \centering
         \includegraphics[width=\textwidth]{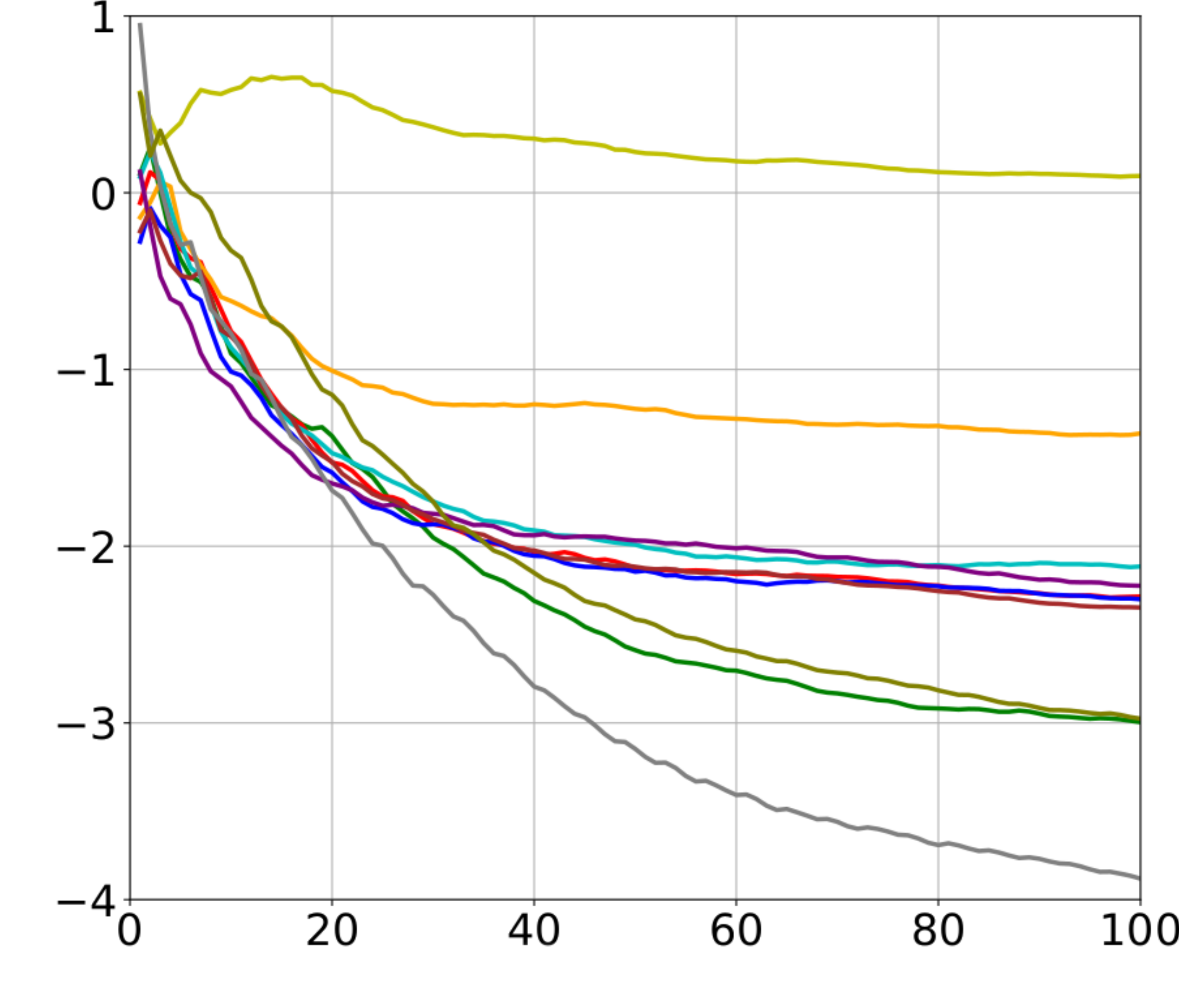} 
         \caption{$\beta$=0.9}
        \Description{The full state space is divided into 10 regions uniformly across the narrowed top right corner.}
     \end{subfigure}
     
        \caption{Average velocity over time for each skill.}
        \label{fig: beta3}     
\end{figure}

\begin{figure}[ht]
     \centering
     \begin{subfigure}[b]{0.25\textwidth}
         \centering
         \includegraphics[width=\textwidth]{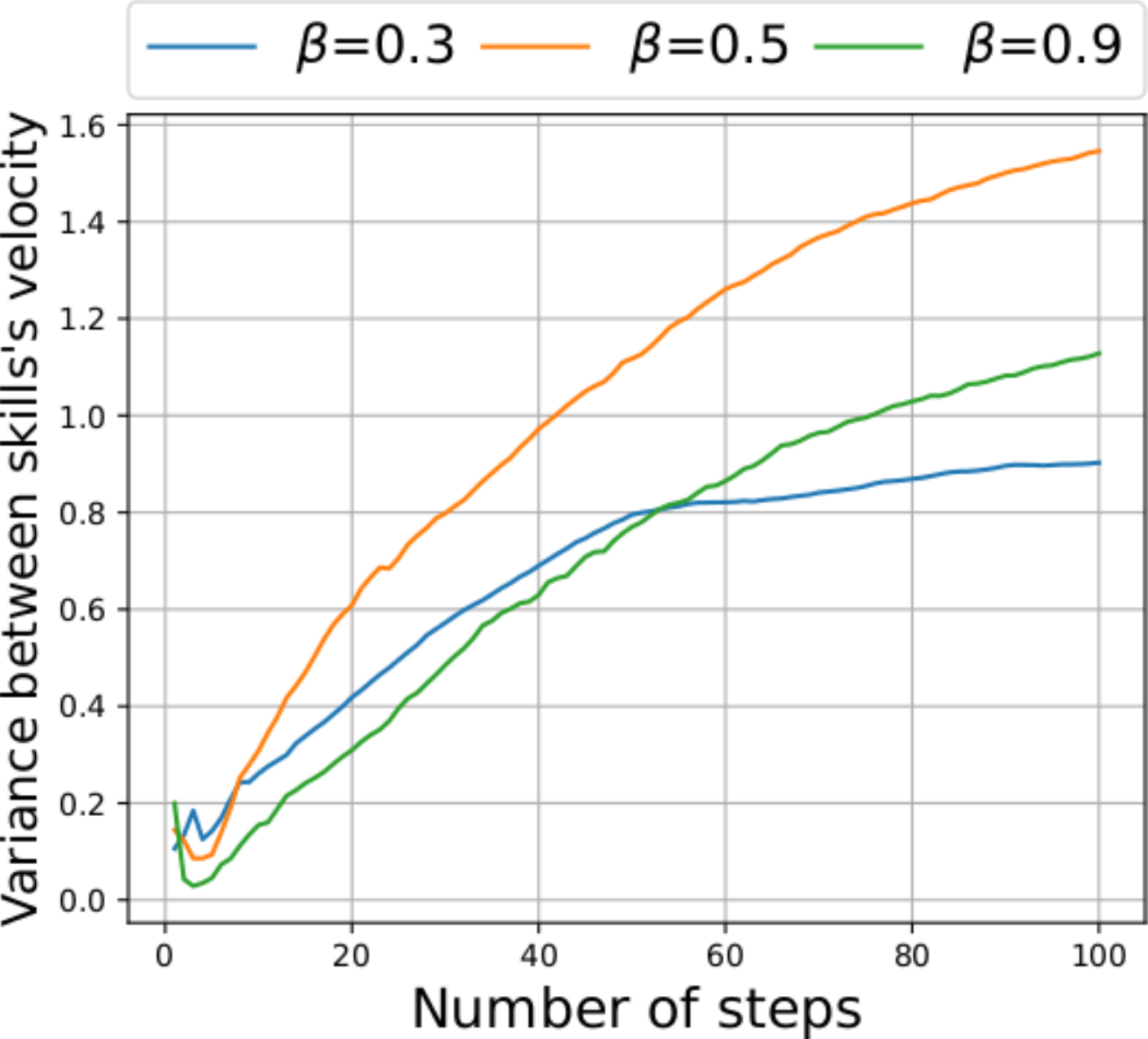} 
     \end{subfigure}
    \caption{Comparaison of the variance between skill's velocity over time for each of the $\beta$ values.}
    \label{fig: beta4}   
\end{figure}

\begin{figure}[ht]
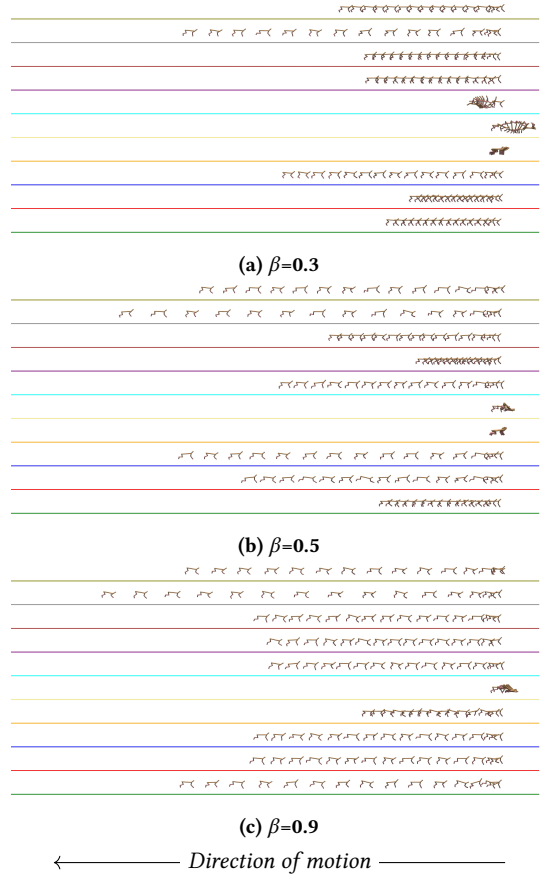

     \centering
     \begin{subfigure}[b]{0.4\textwidth}
         \centering
         \includegraphics[width=\textwidth]{../figures/beta/skills/0.3.pdf} 
         \caption{$\beta$=0.3}
        \Description{The figure shows that the skills learned are moving backward at varying velocity on a short distances.}
     \end{subfigure}
     \begin{subfigure}[b]{0.4\textwidth}
         \centering
         \includegraphics[width=\textwidth]{../figures/beta/skills/0.5.pdf} 
         \caption{$\beta$=0.5}
        \Description{The figure shows that the skills learned are moving backward at varying velocity on different distances.}
     \end{subfigure}
     \begin{subfigure}[b]{0.4\textwidth}
         \centering
         \includegraphics[width=\textwidth]{../figures/beta/skills/0.9.pdf} 
         \caption{$\beta$=0.9}
        \Description{The figure shows that the skills learned are moving backward at varying velocity on a long distance.}
     \end{subfigure}
    \begin{tikzpicture}
          \coordinate (A) at (-3,0);
          \coordinate (B) at ( 3,0);
          \draw[<-] (A) -- (B) node[midway,fill=white] {\emph{Direction of motion}};
    \end{tikzpicture}
        \caption{Modified Half cheetah skills learned with different $\beta$ values.}
        \label{fig: beta2}
\end{figure}

Similarly, in Figure \ref{fig: beta2}, a low $\beta$ results in skills that only cover shorter distances, as these are easier to learn. On the other hand, an agent that exploits the reward (high $\beta$) learns skills that cover larger distances. However, high $\beta$ values may cause the agent to be overly exploitative, leading to a lower diversity of learned skills. This phenomenon is illustrated in Figures \ref{fig: beta3} and \ref{fig: beta4} which show that the variance of velocity across skills is relatively low for both high and low values of $\beta$, while it is the highest for the intermediate value of $\beta=0.5$. Hence, a user favoring a more uniform distribution of skills might choose a more balanced $\beta$ of $0.5$, while one favoring skills more relevant to the task should select a relatively high $\beta$.

\section{Conclusion}

We introduced a novel approach for addressing the issue of under-constrained skill discovery. Our proposed approach, Controlled diversity with preference (CDP) was designed to leverage human feedback to identify human-preferred regions, following which we discovered diverse skills within those regions, thereby ensuring the learning of diverse and desirable skills. In addition, we show that our method can be used to guide exploration towards possible preferred regions. We validated our proposed approach in 2D navigation and Mujoco environments. Empirically, our agents demonstrated the ability to favor the exploration of the preferred regions and to learn diverse skills in these regions. We also empirically studied the effect of the user-controlled hyperparameter $\beta$ to demonstrate its effects on the diversity of learned skills. As such, we believe that our approach presents a way to control the autonomous discovery of skills, while still obtaining safe, aligned and desirable skills.

%%%%%%%%%%%%%%%%%%%%%%%%%%%%%%%%%%%%%%%%%%%%%%%%%%%%%%%%%%%%%%%%%%%%%%%%

%%% The acknowledgments section is defined using the "acks" environment
%%% (rather than an unnumbered section). The use of this environment 
%%% ensures the proper identification of the section in the article 
%%% metadata as well as the consistent spelling of the heading.

%%%%%%%%%%%%%%%%%%%%%%%%%%%%%%%%%%%%%%%%%%%%%%%%%%%%%%%%%%%%%%%%%%%%%%%%

%%% The next two lines define, first, the bibliography style to be 
%%% applied, and, second, the bibliography file to be used.
\balance
\bibliographystyle{ACM-Reference-Format} 
\bibliography{paper/references}
%%%%%%%%%%%%%%%%%%%%%%%%%%%%%%%%%%%%%%%%%%%%%%%%%%%%%%%%%%%%%%%%%%%%%%%%
\end{document}